\documentclass[journal]{IEEEtran}
\ifCLASSINFOpdf
\else
\fi
\usepackage{amsmath, amsthm, amssymb}
\usepackage{algpseudocode}
\usepackage{algorithm}
\usepackage{multirow}
\usepackage{array}
\usepackage{booktabs}
\usepackage{bm}
\usepackage{graphicx}
\usepackage{xcolor}
\usepackage{soul}
\usepackage[utf8]{inputenc}

\usepackage{float}
\usepackage{subcaption}
\usepackage{cases}
\usepackage{mdwmath}
\usepackage{subcaption}
\usepackage{tcolorbox}
\usepackage{authblk}
\usepackage{colortbl}
\usepackage[normalem]{ulem}
\usepackage{siunitx}
\usepackage{hyperref}
\usepackage{dblfloatfix}

\usepackage{textcomp}

\begin{document}
%
\title{Representation Based Regression \\ for Object Distance Estimation}
%
%
%

\author{Mete Ahishali,
        Mehmet Yamac,
        Serkan Kiranyaz,
        and~Moncef~Gabbouj
\thanks{Mete Ahishali, Mehmet Yamac, and Moncef Gabbouj are with the Faculty of Information Technology and Communication Sciences, Tampere University, 33720 Tampere, Finland (email: name.surname@tuni.fi).}
\thanks{Serkan Kiranyaz is with the Department of Electrical Engineering, Qatar University, 2713 Doha, Qatar (email: mkiranyaz@qu.edu.qa).}}

\maketitle

\begin{abstract}
In this study, we propose a novel approach to predict the distances of the detected objects in an observed scene. The proposed approach modifies the recently proposed Convolutional Support Estimator Networks (CSENs). CSENs are designed to compute a direct mapping for the Support Estimation (SE) task in a representation-based classification problem. We further propose and demonstrate that representation-based methods (sparse or collaborative representation) can be used in well-designed regression problems. To the best of our knowledge, this is the first representation-based method proposed for performing a \textit{regression} task by utilizing the modified CSENs; and hence, we name this novel approach as \textit{Representation-based Regression (RbR)}. The initial version of CSENs has a proxy mapping stage (i.e., a coarse estimation for the support set) that is required for the input. In this study, we improve the CSEN model by proposing Compressive Learning CSEN (CL-CSEN) that has the ability to jointly optimize the so-called proxy mapping stage along with convolutional layers. The experimental evaluations using the KITTI 3D Object Detection distance estimation dataset show that the proposed method can achieve a significantly improved distance estimation performance over all competing methods. Finally, the software implementations of the methods are publicly shared at https://github.com/meteahishali/CSENDistance.
\end{abstract}


%
\IEEEpeerreviewmaketitle

\section{Introduction}

\IEEEPARstart{D}{istance} estimation has been a crucial task since its application plays a vital role in many autonomous frameworks, e.g., autonomous driving, unmanned aerial vehicles, and robotics. One can estimate the object specific distance from the depth scene produced by depth sensors such as LiDAR or utilizing such methods that use only visual information. Naturally, the latter is preferable because of the extra cost of LiDAR. Moreover, even though the LiDAR sensor can operate under varying weather conditions, it has a limited coverage area such as 5\% of the image space \cite{plug-and-play}. Hence, there have been various methods \cite{plug-and-play,disnet,iccv,vid2depth,struct2depth,stereo,svr} that focus on developing computer vision solutions for the depth estimation including supervised and unsupervised approaches. For example, the method in \cite{stereo} has utilized multiple cameras to compensate for the lack of sensors. On the other hand, the need for multiple cameras and processing costs are disadvantages of a stereo-camera based depth estimation method. Thus, several methods have studied monocular depth estimation \cite{vid2depth,struct2depth}, and they have revealed that by following recent trends in neural networks, i.e., fully convolutional neural networks, depth estimation performance with a single RGB image can be comparable enough with a stereo-camera based approaches. As unsupervised learning strategies, studies in \cite{vid2depth,struct2depth}, propose to learn depth information from structural changes within consequent frames. Additionally, besides using the visual data alone, a hybrid approach combining and utilizing both visual and sensor data can be another alternative for enhancing the noisy or erroneous depth predictions. For example, the authors claim in \cite{plug-and-play} that their method can be integrated into various learning-based methods that use visual information, and it can improve the performance of the methods by sparse LiDAR measurements.

Nevertheless, the aforementioned methods except \cite{disnet,iccv}, and \cite{svr} have focused on producing dense depth maps which means computing a heat-map that gives a sense of relative depth distance information in an observed scene. On the other hand, the necessity of dense depth maps varies among applications, i.e., in an autonomous driving application, the distance information of the objects is more desirable than providing the depth map of the scene. There are only a few studies, \cite{disnet,iccv,svr} that propose object distance estimation for the objects in an observed scene.

The pioneer study \cite{disnet} of object distance estimation on land proposes a two-tiers methodology: i) first, detection of the location, and then the classification of an object. ii) extraction of the related features such as the bounding box information (the width, the height, etc.), and class-specific ones (such as predefined average length of detected class). (iii) Finally, using a Multi-Layer Perceptron (MLP) to predict the camera distance of the bounding box in meters. However, their approach directly depends on the performance of object classification, while a misclassification of the given ROI may lead to a complete failure in distance estimation. Similarly, the study in \cite{svr} proposes to use only the geometric information of the bounding box as features and train a Support Vector Regressor (SVR). On the other hand, in \cite{iccv}, a Convolutional Neural Network (CNN) is used to extract representative features and these features have then been used for the regression and the classification tasks by two MLPs to predict the distance of the object and its category. Since the overall framework in \cite{iccv} is trained jointly by combining the classification and regression losses, the categorical information of the objects has boosted the estimation of the distance.

Overall, comparing recent improvements in the dense depth estimation \cite{vid2depth,struct2depth,plug-and-play}, there is a lack of existing research focusing on object distance estimation. In this study, we believe that the importance of object distance estimation is obvious as the number of recent advances in the \textit{state-of-the-art} object detectors has been growing and further analysis over these objects can provide better assistance to autonomous systems.

Deep Learning approaches with the recent advances in Convolutional Neural Networks (CNNs) have provided \textit{state-of-the-art} performance levels in various computer vision tasks such as object detection, image recognition, and image segmentation. To achieve such performance levels, the deep learning-based approaches require a massive training dataset. On the other hand, the proposed solution for the object distance estimation task should be suitable to work with relatively small-scale annotated data. For example, one can compare KITTI 3D Object Detection \cite{kitti} dataset having annotated 7481 scenes with Imagenet \cite{imagenet} having over a million samples.

To address this need, in this study, we first formulate the distance estimation problem as a multi-class classification task by quantizing the distance in meters and use representation-based classification techniques including two categories: Sparse Representation-based Classification (SRC) and Collaborative Representation-based Classification (CRC). The approaches for SRC \cite{SRC1, SRC2} and CRC \cite{collaborative} are well suited for the limited data and they are commonly used for the classification in the existing studies as follows. A representative dictionary $\mathbf{D}$ is constructed by grouping training samples column-wise. In the inference phase, a test sample $\textbf{y}$ will be attempted to be represented by the linear combination of the atoms of the formed dictionary $\mathbf{D}$, i.e., solving $\mathbf{y} = \mathbf{Dx}$ for $\mathbf{x}$ where $\mathbf{x}$ is a vector of representation coefficients. Accordingly, in SRC methods, it is aimed to find a sparse $\mathbf{\hat{x}}$ (just have enough non-zero components so that the query sample is represented with a small error margin). Alternatively, in CRC, the least-square sense solution is applied, i.e., $\hat{\mathbf{x}} = \left( \mathbf{D}^T\mathbf{D} + \lambda \mathbf{I} \right)^{-1}\mathbf{D}^T\mathbf{y}$, where $\lambda$ is the regularization parameter. Overall, the same motivation is valid for both categories: the atoms having higher estimated representation coefficient values, $\hat{\mathbf{x}}$, are likely to have the same class label with the query sample $\mathbf{y}$. It has been observed in \cite{collaborative} that the CRC approach has provided marginally reduced classification performance compared to SRC methods. However, the computational complexity of the methods that rely on SRC is significant considering that they require iterative computations to solve the problem.

In this study, we propose the following approach of using a representation-based scheme in the object distance estimation task. First, the cropped objects are resized to have fixed size images for each object, and then their corresponding features are obtained by using pre-trained networks DenseNet-121 \cite{DenseNet}, VGG19 \cite{vgg19}, and ResNet-50 \cite{resnet50} over the ImageNet dataset. Next, a dictionary is created with atoms of relative features that are from the classes obtained by discretizing the distances of the object. Finally, a representation-based classification method is applied to detect the class which will correspond to the discretized distance of the query object. The main advantage of the proposed approach is that the categorical information of the object is not used in the distance estimation unlike the methods in \cite{disnet, iccv}; hence the classification performance of a single-stage object detector does not affect the distance prediction performance. 

As an alternative approach, we propose to consider this as a regression problem. In order to make a direct distance estimation without discretization during the inference phase, we modify Convolutional Support Estimator Network (CSEN) that was originally proposed as a representation-based classifier in \cite{csen}. The CSEN approach combines the conventional representation-based classification technique with the learning-based approach involving CNNs. We define the task of Support Estimation (SE) to estimate locations of the non-zero components of $\mathbf{x}$. Indeed, the support of the non-zero coefficient forms sufficient information to obtain the class of the query sample. The previous works \cite{csen, csen-early-covid, csen-covid} have shown that CSENs provide \textit{state-of-the-art} classification performance levels and their computational complexity are insignificant since they can directly map the support set of the query sample. Moreover, they are well-suited for limited annotated data since they do not have the tendency to \textit{overfit} due to their compact structures. Up to date, the CSEN approach has never been designed and evaluated for a regression task. In this study, we show that using the modified CSEN configuration, it is possible to perform a regression task that is henceforth called as \textit{Representation-based Regression (RbR)}. Finally, we propose further improvements over the CSEN framework. The initial CSEN version \cite{csen} has required the so-called proxy, $\tilde{\mathbf{x}}$, estimation based on the least-square solution, i.e., $\tilde{\mathbf{x}} = \left( \mathbf{D}^T\mathbf{D} + \lambda \mathbf{I} \right)^{-1}\mathbf{D}^T\mathbf{y}$. In this study, we propose an end-to-end learning, the so-called Compressive Learning CSEN (CL-CSEN) framework that \textit{jointly} optimizes the proxy mapping and SE estimation.

Overall, the novel and significant contributions of this study can be summarized as follows:
\begin{itemize}
    \item Representation-based classification approaches are used in an object-specific distance estimation task for the first time.
    \item To the best of our knowledge, this is the first study that formulates a regression task in the form of representation-based estimation approach. The proposed methodology is henceforth named as \textit{Representation-based Regression (RbR)}.
    \item With the proposed approach, the \textit{state-of-the-art} performance level is achieved using compact configurations and a non-iterative SE. This does not only enables an accurate estimation with a limited number of annotated data, it further yields an elegant efficiency in terms of computational complexity.
    \item The improved framework with CL-CSEN enables the joint optimization of CSEN framework with the denoiser matrix $\mathbf{B}$ in order to be used in proxy mapping, i.e., $\tilde{\mathbf{x}} = \mathbf{By}$.
    \item Finally, contrary to the depth estimation, there is a limited number of studies proposed for the object-specific distance estimation including \cite{disnet,iccv,svr}. Furthermore, the studies in \cite{disnet,iccv} require additional information besides a single RGB image such as the object class information and camera projection matrix.
\end{itemize}

Our experimental evaluations over the KITTI benchmark dataset \cite{kitti} show that the distance estimation performance with the proposed CSEN approach outperforms all competing methods, i.e., the competing distance estimator SVR \cite{svr} and alternative representation-based approaches including CRC \cite{collaborative} and SRC approaches \cite{SRC1, SRC2}. Moreover, although the direct comparison of the proposed approach is not fair against the method in \cite{iccv} due to the reasoning mentioned earlier, the proposed approach still outperforms \cite{iccv} considering their reported performance metrics.

The rest of the paper is structured as the following: the theoretical background and the prior art will be presented in Section \ref{background}. Then, the proposed object distance estimation with CSEN and CL-CSEN will be detailed in Section \ref{proposed-methodology}. Next, the experimental evaluations over the KITTI dataset are presented in Section \ref{results}. Finally, concluding remarks will be drawn in Section \ref{conclusion}.

\section{Background and Prior Art}
\label{background}

In this section, we shall first provide a brief background of sparse representation, then, discuss the representation-based classification theory including SRC and CRC methods.

The following notations and terms are defined in this study. For a vector, $\mathbf{x} \in \mathbb{R}^n$, the $\ell_p$-norm is $\left \| \mathbf{x} \right \|_{\ell_p^n} =   \left (  \sum_{i=1}^n \left \vert x_i \right \vert^p \right )^{1/p}$ where $p \geq 1$, whereas the $\ell_0$-norm and $\ell_{\infty}$-norm are defined as $\left \| \mathbf{x} \right \|_{\ell_0^n} = \lim_{p \to 0} \sum_{i=1}^n \left \vert x_i \right \vert^p = \# \{ j: x_j \neq 0 \}$ and $\left \| \mathbf{x} \right \|_{\ell_{\infty}^n} =  \max_{i=1,...,n} \left (   \left | x_i \right | \right )$ for the vector $\mathbf{x}$, respectively. Let a signal $\mathbf{s}$ is sparsely represented in a domain $ \mathbf{ \Phi}$ such that $\mathbf{s}= \mathbf{ \Phi}\ \mathbf{x}$ where $\left \|  \mathbf{x} \right \|_0 \leq k$, then it is said that the signal $\mathbf{s}$ is strictly $k$-sparse since it can be represented using less than $k+1$ non-zero coefficients in a proper domain. That is to say, it is possible to represent the signal $\mathbf{s}$ with only a few basis vectors in a proper domain $\mathbf{ \Phi}$. The sparse support set $\Lambda$ is then a set that contains locations of these non-zero coefficients of $\mathbf{x}$ such that $\Lambda := \left \{ i:  x_i \neq 0 \right \}$ and $\Lambda \subset \{1,2,3,...,n \}$.

Let $\mathbf{A}$ is a subspace for the signal $\mathbf{s}$ such that $\mathbf{y} = \mathbf{A} \mathbf{s}$. Accordingly, a signal $\mathbf{y}$ can be projected to the subspace $\mathbf{A}$ as follows:
\begin{equation}
     \mathbf{y} = \mathbf{A} \mathbf{s} = \mathbf{A}\mathbf{ \Phi} \mathbf{x} =\mathbf{ D} \mathbf{x}
     \label{CS},
\end{equation}
where $\mathbf{A} \in \mathbb{R}^{m \times d}$ is called compression matrix, $\mathbf{D} \in \mathbb{R}^{m \times n}$ is the equivalent dictionary, and $m < < n$; and hence the corresponding system is underdetermined. We necessitate a priori information regarding the unknown $\mathbf{x}$ to solve such an ill-posed problem in \eqref{CS} since it is non-uniquely solvable. The study in \cite{spark} has shown that at least $k$-sparse signal pairs in a sparsifying basis $\mathbf{ \Phi}$ are distinguishable in the dictionary $\mathbf{D}$ if $\mathbf{D}$ satisfies some properties. Consequently, it immediately indicates that the below solution is unique,
\begin{equation}
\min_\mathbf{x} ~ \left \| \mathbf{x }\right \|_{0}~ \text{subject to}~ \mathbf{D} \mathbf{x} = \mathbf{y}, \label{sparse_rep}
\end{equation}
if $\left \| \mathbf{x }\right \|_{0} \leq k$, $m \geq 2k$, and the minimum number of linearly independent columns in $\mathbf{D}$ is also greater than $2k$ \cite{spark}.

On the other hand, the solution of \eqref{sparse_rep} is NP hard and the problem is non-convex. Fortunately, we can relax the optimization problem, $\ell_0$-minimization, to its closest norm based problem defined as Basis Pursuit \cite{BP} that is $\ell_1$-norm:
\begin{equation}
     \min_\mathbf{x} \left \| \mathbf{x} \right \|_1 ~s.t. ~ \mathbf{x} \in \mho \left (\mathbf{ y} \right ) \label{Eq:l1}
\end{equation}
where $\mho \left ( \mathbf{y} \right ) = \left \{ \mathbf{x}: \mathbf{D} \mathbf{x}=\mathbf{y} \right \}$. The equivalent to the one of the sparse representation problem \eqref{sparse_rep}, but more tractable solution can be achieved by solving $\ell_1$-minimization defined in \eqref{Eq:l1} under some conditions such as $m>k(log(n/k))$ and $\mathbf{D}$ satisfied Restricted Isometry Property \cite{candesRIP}.

\subsection{Generic Sparse Support Estimation (SE)}

Support estimation can be defined as finding the non-zero locations of a corresponding sparse signal. Indeed, in many practical problems, the full signal recovery; the recovery of the signal magnitude, sign, and support set, may not be necessary. For example, in a representation-based classification problem as in \cite{SRC1,SRC2,collaborative}, estimating the locations of non-zero elements in $\mathbf{x}$ so-called the support set, $\Lambda$, is enough to determine the corresponding class. Let the linear feed-forward model be $\mathbf{y}= \mathbf{Dx} +\mathbf{z}$ with an additive noise $\mathbf{z}$, then a support estimator $\mathcal{E}(.)$ will estimate the indices of the non-zero elements given $\mathbf{D}$ and $\mathbf{y}$, i.e.,
\begin{equation}
   \hat{\Lambda} =  \mathcal{E}\left (\mathbf{y},\mathbf{D} \right ) 
\end{equation}

The works in the literature targeting SE from $y$, i.e., $\hat{\Lambda} = \mathcal{E}\left (\mathbf{y},\mathbf{D} \right ) $, are based on by first applying a signal recovery method then applying component-wise thresholding over the estimated signal, $\hat{\mathbf{x}}$, to compute $\hat{\Lambda}$. Accordingly, they can be divided into three categories depending on their reconstruction schemes: (i) estimators that are based on $\ell_1$-minimization, (ii) least-square sense approximate methods such as LMSEE \cite{AMP-partial}, $\hat{\mathbf{x}}^{LMMSE} = \left ( \mathbf{D}^T \mathbf{D} + \lambda \mathbf{I}_{n \times n} \right )^{-1} \mathbf{D}^T \mathbf{ y}$ and Maximum Correlation (MC) \cite{MaximumCorrelation}, $\hat{\mathbf{x}}^{MC} = \mathbf{D}^T \mathbf{ y}$, and (iii) Deep Neural Networks.

The approaches in (i) work in an iterative manner and they are computationally costly; and hence, not efficient if the aim is to only recover support information. The methods in (ii) are non-iterative and direct approaches but their performances are limited compared to the previous ones (see \cite{AMP-partial} for a detailed discussion). Finally, deep learning-based approaches \cite{Lamp} in the group (iii) target a direct mapping for the signal reconstruction task. However, the major concern is that the signal reconstruction task is harder than SE, and it requires deep networks having complex architectures with millions of parameters to enable a direct mapping. This further requires a massive size of training data for a proper generalization. Furthermore, these deep unfolding networks \cite{Lamp} consist of dense layers making them computationally intensive and more sensitive to the additional noises \cite{csen}. As a remedy, our recent approach, CSEN \cite{csen}, which can perform direct SE without first applying signal recovery, provides an alternative and computationally efficient solution. The compact design of CSEN enables elegant performance even with small-scale training data. Moreover, compared to deep networks with dense layers, CSEN with convolution layers provide robust SE in noisy cases. For a more detailed analysis, the readers are referred to \cite{csen} in which we compare the performances of the traditional support estimators with the proposed CSEN approach and we address the major limitations and drawbacks with the classical support estimator methods.

\begin{figure*}[]
\centering
  \includegraphics[width=1\linewidth]{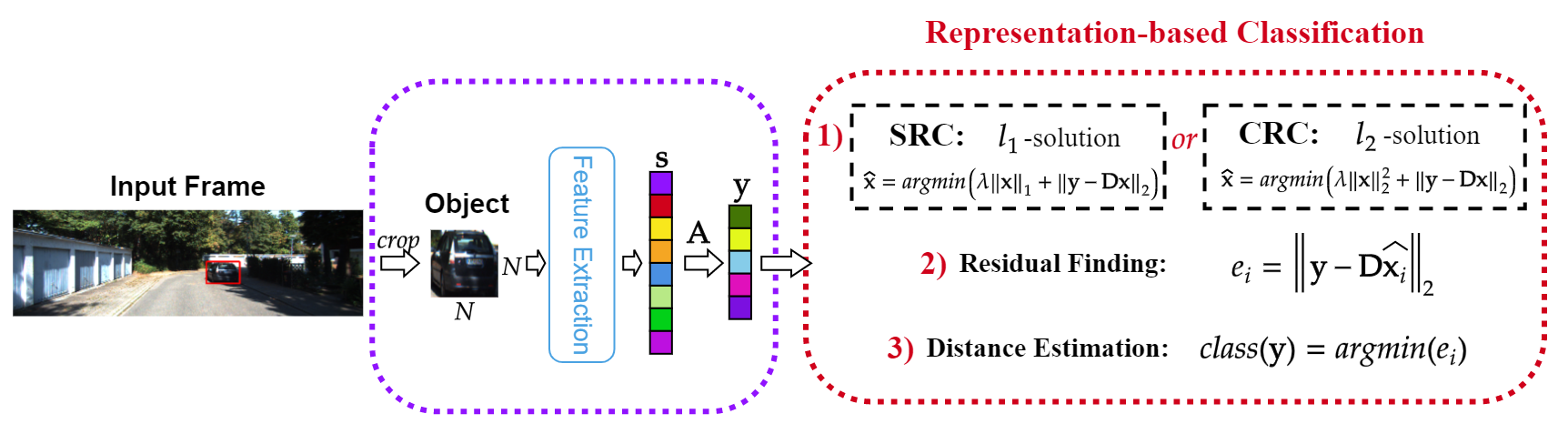}
  \caption{The proposed framework for the object distance estimation is based on representation-based classification methodologies including Sparse Representation-based Classification (SRC) and Collaborative Representation-based Classification (CRC). The output class estimation yields the quantized estimated distance.}
  \label{src-crc}
\end{figure*}

\subsection{Representation-based Classification}
\label{src-crc-approaches}

As discussed earlier, in representation-based classification task, predicting the locations of the non-zero elements in $\mathbf{x}$ is more important than computing the exact values. In the following sub-sections, we shall provide a brief explanation about how SRC and CRC methods perform classification. Basically, SRC methods are in the aforementioned first group of support estimators, whereas the CRC method belongs to the second group.

\subsubsection{Sparse Representation-based Classification} When a test sample $\mathbf{y}$ is introduced, the query sample $\mathbf{y}$ is tried to be represented as a linear combination of the atoms of the dictionary $\mathbf{D}$. In general, SRC methods estimate sparse representation coefficients $\hat{\mathbf{x}}$ that only a few non-zero coefficients exist to represent the query sample. We expect that these active components of $\hat{\mathbf{x}}$ will correspond to the samples having the same label as the test sample. There are many existing studies that utilize the SRC approach for various classification tasks such as face recognition \cite{SRC2}, coronavirus disease 2019 (COVID-19) recognition \cite{csen-covid}, early COVID-19 detection \cite{csen-early-covid}, human action recognition \cite{human-action}, and hyper-spectral image classification \cite{hyperspecral}.

In the previous discussion regarding \eqref{sparse_rep} and \eqref{Eq:l1}, the statements were valid for the exactly $k$-sparse signal pairs, whereas in practice, the signal $\mathbf{x}$ may not be exactly $k$-sparse due to the modeling errors or noise in the data. Consequently, given the measurement with the additive noise: $\mathbf{y} =\mathbf{D} \mathbf{x} + \mathbf{z}$, the exact recovery of the signal is unfeasible. However, the stable signal recovery is still possible in which the stable recovery refers that $\hat{\mathbf{x}}$ obeys $\left \| \mathbf{x }- \hat{\mathbf{x}} \right \| \leq \kappa \left \| \mathbf{z} \right \|$ hold for the estimated sparse signal $\hat{\mathbf{x}}$, where $\kappa$ is a relatively small constant. For instance, it is provided in \cite{lasso-stable} that using the following so-called Lasso formulation:
\begin{equation}
\label{lasso}
    \min_\mathbf{x}  \left \{ \left \|  \mathbf{D}\mathbf{x}-\mathbf{y} \right \|_2^2 + \lambda \left \|\mathbf{ x} \right \|_1  \right \}    
\end{equation}
the partial recovery of the sparse $\mathbf{x}$ is achievable. Correspondingly, it is also proven in \cite{lasso-stable} that $\ell_1$ solution can still provide exact computing of $\mathbf{x}$ in noise-free cases.

In \cite{SRC2}, a four-step approach is proposed instead of using \eqref{lasso} directly: i) normalize all the atoms in $\mathbf{D}$ and $\mathbf{y}$ to have unit $\ell_2$-norm, ii) apply the signal reconstruction step: $\hat{\mathbf{x}} = \arg \min_{\mathbf{x}} \left \|\mathbf{ x} \right \|_1 \text{s.t.} \left \| \mathbf{y} - \mathbf{D} \mathbf{x} \right \|_2 $, (iii) residual finding: $\mathbf{e_i} = \left \| \mathbf{y} - \mathbf{D_i}  \mathbf{\hat{x}_i} \right \|_2$, where $\mathbf{\hat{x}_i}$ is the estimated coefficients corresponding the class $i$, (iv) estimated label: $\text{Class}\left ( \mathbf{y} \right ) = \arg \min \left ( \mathbf{e_i} \right )$. Although this four-step solution introduces an additional residual finding step, it provides performance improvements over direct SE with \eqref{lasso} since the samples from different classes are actually correlated as in real life. The other SRC techniques in \cite{human-action, hyperspecral} have followed similar approaches with \cite{SRC2}.

\subsubsection{Collaborative Representation-based Classification}

The study in \cite{collaborative} proposes to follow $\ell_2$-minimization instead of $\ell_1$-minimization in \eqref{lasso} as follows:
\begin{equation}
    \label{l2}
    \mathbf{\hat{x} }= \arg \min_{\mathbf{x}} \left \{ \left \| \mathbf{y }- \mathbf{D}\mathbf{x }\right \|_2^2  + \lambda \left \| \mathbf{x} \right \|_2^2   \right \}
\end{equation}
Hence, they form the CRC approach in \cite{collaborative} by changing the second step of the four-step solution in \cite{SRC2} with the following closed-form solution: $\hat{\mathbf{x}} = \left ( \mathbf{D}^T  \mathbf{D} + \lambda \mathbf{I}_{n \times n}  \right )^{-1} \mathbf{D}^T  \mathbf{ y}$. The motivation is that for a given a query signal $\mathbf{y}$ or vectorized image, the computed $\hat{\mathbf{x}}$ should have minimum energy with relatively small coefficients that correspond to samples in the dictionary $\mathbf{D}$ from the same class with the query $\mathbf{y}$. Hence, due to the least-square sense minimization technique, a collaborative representation is sought between the atoms of the dictionary. In representation-based classification scheme, the dictionary $\mathbf{D}$ mostly fails to satisfy the defined exact or robust recovery properties due to the correlation between samples. It is indeed discussed in \cite{collaborative} that if formulating the problem with the collaborative representation operates the classification rather than the sparse representation.

It is reported that the followed $\ell_2$-minimization based solution provides especially high classification performances for a high compression ratio that is defined as $m/d$. In those cases, the CRC approach can even produce comparable or better classification results comparing with SRC. Note the fact that the CRC approach is considerably faster due to the presence of the closed-form solution in \eqref{l2}.

\section{The Proposed Methodology}
\label{proposed-methodology}

In the sequel, we will introduce the feature extraction procedure and the framework about using classical representation-based classification methods: SRC and CRC on the distance estimation task with the quantization. Then, the proposed CSEN based regression approach will be presented to directly predict the distance information without the quantization in the inference. Finally, a novel CL-CSEN framework will be introduced that is specifically designed to jointly optimize the denoiser and the regression parts of the CSEN during the training phase.

\subsection{Estimation via Representation-based Classification}
\label{rpc}

The representative dictionary that is needed to form in the representation-based classification methods can be formed by vectorized samples. However, we have revealed in \cite{csen} that for some cases, the atoms of the collected dictionary for a representation-based classification method are not representative enough if they are formed by directly putting the vectorized raw images. Hence, in the regression task as well, we propose to use a pre-trained CNN to produce more representative information.

The selected pre-trained models for this feature extraction procedure are DenseNet-121 \cite{DenseNet}, VGG19 \cite{vgg19}, and ResNet-50 \cite{resnet50} that are trained over the ImageNet dataset with more than one million images:
\begin{itemize}
    \item DenseNet-121 is a fully connected convolutional network: an $L$-layer DenseNet-121 has a total of $L(L+1)/2$ connections whereas the corresponding traditional version of the convolutional network would have only $L$ connections.
    \item VGG19 is a deep neural network consisting of convolutional and fully connected (dense) layers (as a generic CNN structure) without any skip-connections.
    \item ResNet-50 is based on residual learning having skip-connections between every other layer in the network.
\end{itemize}
Overall, DenseNet-121 and ResNet-50 are in the form of convolutional layers consisting of only convolutional layers except for the output layer, whereas VGG19 has multiple fully connected layers.

\begin{figure}[]
  \includegraphics[width=1\linewidth]{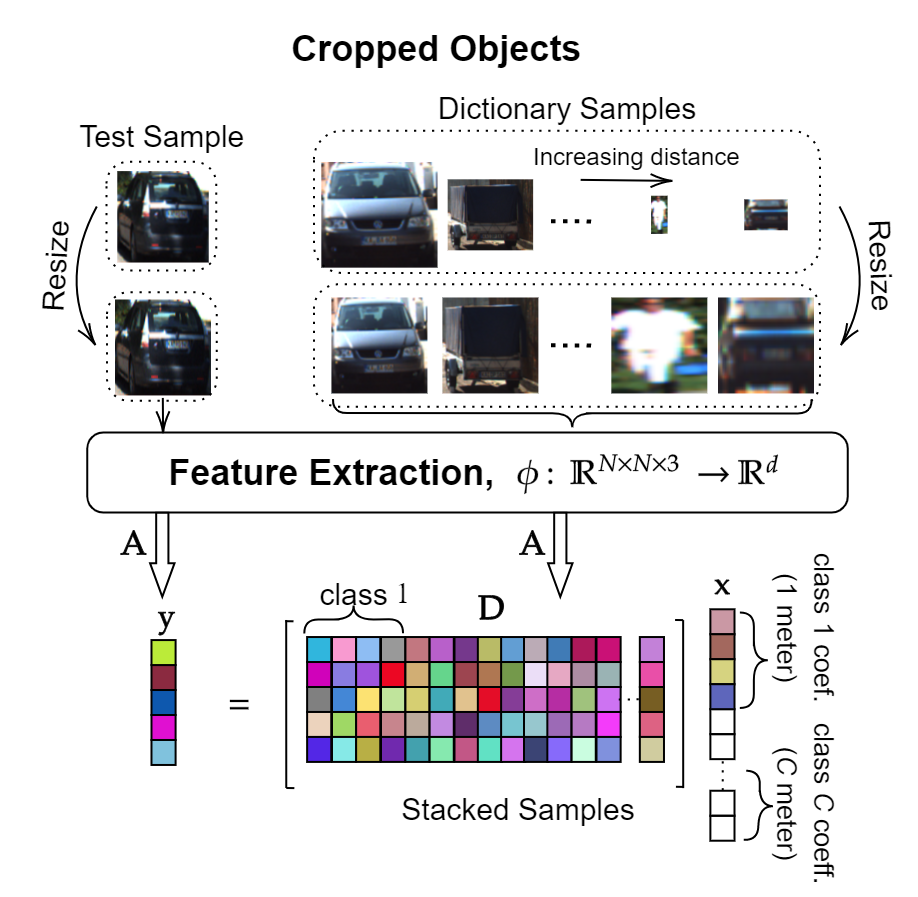}
  \caption{To form the representative dictionary $\mathbf{D}$, samples are collected with the increasing order of the distances. Then, they are resized and fed to the feature extractor. Next, after additional dimensional reduction operation with the matrix $\mathbf{A}$, they are stacked in such a way that the first-class category corresponds to 1m and the $C^{\text{th}}$ class to $C$ meters.}
  \label{dictionary_design_src_crc}
\end{figure}

Accordingly, we compose the features before the last convolutional layers of DenseNet-121 and ResNet-50 and before the fully connected layers of VGG19. Then, the collected multiple feature maps are flattened by applying global max-pooling operation. Consequently, the described feature extraction procedure provides the mapping $\phi: \mathbb{R}^{N \times N \times 3} \rightarrow \mathbb{R}^{d}$ to produce a feature vector, $\mathbf{s}_i=\phi\left ( \mathbf{I}_i \right )$, where $\mathbf{I}_i$ is the $i^{\text{th}}$ object cropped from the observed frame and resized to a predetermined size i.e., $N \times N$ as demonstrated in Fig. \ref{src-crc}. The feature vector dimension $d=1024$, $512$, and $2048$ for DenseNet-121, VGG19, and ResNet-50, respectively. The composed features for $m$ number of objects are collected column-wise to have the matrix, $\mathbf{ \Phi} \in \mathbb{R}^{d \times m}$. The representative dictionary $\mathbf{D}$ is then formed as $\mathbf{D} = \mathbf{A \Phi}$ using the compression matrix $\mathbf{A} \in \mathbb{R}^{m \times d}$ as PCA. When forming the dictionary, we quantize the distances to interpret the regression problem as a classification problem. For example, let the desired sensitivity is selected as 1m, then there would be 60-classes for a distance estimation task for the range of 1 - 60 meters. This dictionary formation procedure is illustrated in Fig. \ref{dictionary_design_src_crc}. Apparently, the distance information in the extracted features comes from the resolution of the cropped input images since the distant objects tend to have blurry appearances due to the rescaling small-scale distant objects as observed in Fig. \ref{dictionary_design_src_crc}. Next, representation-based classification approaches with SRC and CRC can be used to predict the class which will correspond to the quantized distance. As illustrated in Fig. \ref{src-crc}, the aforementioned four-step approach in Section \ref{background} is used in the framework including the residual finding step.

\subsection{The Proposed Representation-based Regression (RbR) with CSENs}

\begin{figure*}[ht]
\centering
  \includegraphics[width=1\linewidth]{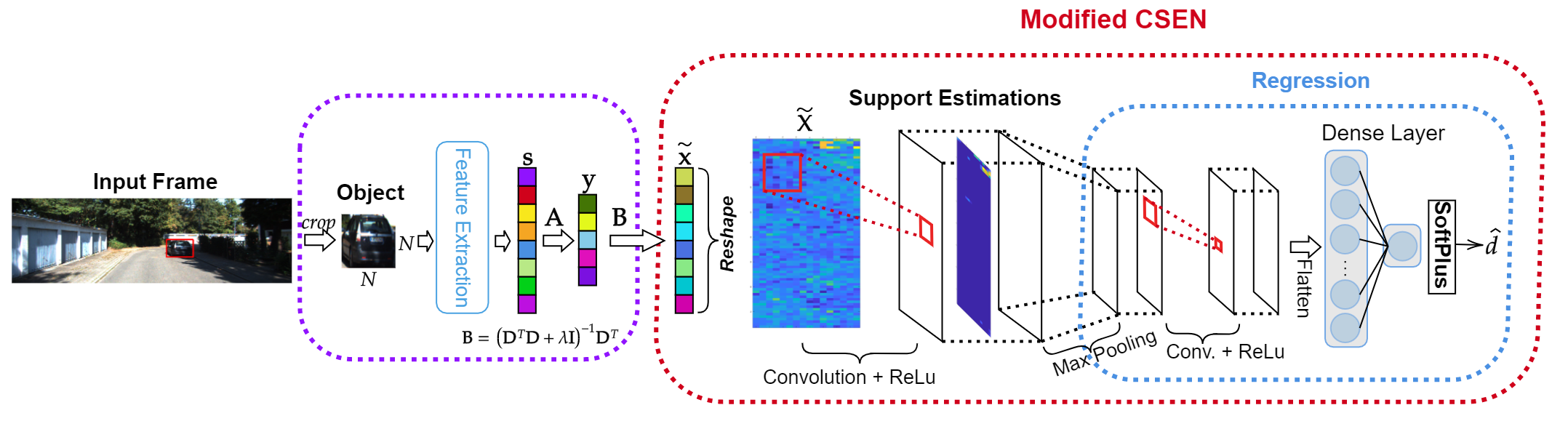}
  \caption{The proposed framework for the object distance estimation based on Convolutional Support Estimator Networks (CSEN). The modified CSEN performs regression over the estimated support sets using the reshaped proxy signal $\mathbf{\Tilde{x}} = \mathbf{By}$ where $\mathbf{B}=\left( \mathbf{D}^T\mathbf{D}+\lambda\mathbf{I} \right)^{-1}\mathbf{D}^T$.}
  \label{csen-diagram}
\end{figure*}

With the proposed approach, it is possible to produce exact estimates instead of quantized distances during the inference. Hence, to the best of our knowledge, as the first time in the literature, we are introducing the utilization of a representative dictionary for a complete regression task. Accordingly, the proposed approach that will be detailed next is called \textit{Representation-based Regression (RbR)}.

Since the traditional approaches first fully reconstruct the signal before the actual SE task, the performance of the SE becomes highly dependent on the performance of the signal recovery. As discussed earlier, the signal reconstruction is not guaranteed if the required sparsity of $\mathbf{x}$ does not hold due to practical reasons such as the presence of significant noise or high correlation between samples as observed in some classification problems such as face recognition. Nevertheless, it is still possible to recover  $\Lambda$ fully \cite{exact4,SE1,exact1, Volkan} or partially \cite{Volkan,SE3, partial2}. With this motivation, we aim to learn a direct mapping to the corresponding support set $\mathbf{\hat{\Lambda}}$ for a given query sample $\mathbf{y}$.

The proposed SE follows a compact architecture that also maximizes the performance with a minimum number of annotated data. To this end, compact CSENs used in the previous work \cite{csen} have been modified to enable regression in the distance estimation task. The proposed modified CSEN inherits the same capabilities and advantages as the previous study in \cite{csen}. The network was a fully convolutional network consisting of only convolutional layers. In this study, we keep this strategy for SE as well and use MLP only for the regression over the predicted support sets. One may consider using an MLP-like structure for SE as in \cite{Lamp} which is originally proposed for SR. However, the followed topology in CSENs brings several advantages over MLPs as proven in \cite{csen}: low computational complexity, robustness to the noise, and learning capability with a limited amount of training data thanks to the compact structure of CSEN and significantly less number of parameters compared to the MLP.

A CSEN network is designed to produce a binary mask $\mathbf{v} \in \left \{ 0,1 \right \}^n$:
\begin{subnumcases}
{v_i =}
   1    \hfill & \text{ if $ i \in \Lambda $ } \\
   0 & \text{ else }. 
\end{subnumcases}
The estimated support set would be $\hat{\Lambda} = \left \{  i \in  \left \{  1,2,..,n\right \} : \hat{v}_i =1   \right \}$. Thus, in \cite{csen}, the CSEN network, which provides $\mathcal{P} \left ( \mathbf{y}, \mathbf{D} \right ): \mathbb{R}^n  \mapsto \left [ 0,1 \right ]^n$ mapping, produces a probability vector $\mathbf{p}$ of each index to be counted as a support. The final $\hat{\Lambda}$ is then obtained by thresholding $\mathbf{p}$ with a fixed threshold.

During the training phase, CSEN takes $\mathbf{\Tilde{x}}$ as the input and produces $\hat{\mathbf{v}}$ as the SE, where $\mathbf{\hat{v}}, \mathbf{\Tilde{x}} \in \mathbb{R}^n$; hence the learned transformation would be  $\hat{\mathbf{v}} \leftarrow    \mathcal{P} \left ( \mathbf{\Tilde{x}} \right )$. Here, the input of CSEN is a rough estimation and it is called \textit{proxy}. The proxy $\mathbf{\Tilde{x}}$ can be the Maximum Correlation $\mathbf{\Tilde{x}=D^Ty}$ or LMMSE \cite{AMP-partial} $\left ( \mathbf{D}^T \mathbf{D} + \lambda \mathbf{I} \right )^{-1} \mathbf{D}^T\mathbf{y}$. The input proxy $\mathbf{\tilde{x}}$ is then reshaped to a 2-D plane and convolved with the weight kernels $\{\mathbf{w}_1^1, \mathbf{w}_1^2, ... , \mathbf{w}_1^N\}$. After the addition of biases $\{b_1^1, b_1^2, ... , b_1^N\}$, the feature tensor $\mathbf{F}_1=\{ \mathbf{f}_1^1, \mathbf{f}_1^2, ..., \mathbf{f}_1^N \}$ in the first hidden layer with $N$ number of weight kernels is formed:
\begin{equation}
\mathbf{F}_1 = \{\text{S}(\text{ReLu}(b_1^i + \mathbf{w}_1^i * \Tilde{\mathbf{x}}))\}_{i=1}^{N},
\end{equation}
where $\text{S}(.)$ is the down- or up-sampling operation and $\text{ReLu}(x) = \text{max}(0, x)$. This is illustrated in Fig. \ref{csen-diagram}. At the layer $l$, the $k^{\rm {th}}$ feature can be defined as follows:
\begin{equation}
    \mathbf{f}_l^k = \textsc{S}(\textsc{ReLu}(b_l^k + \sum_{i=1}^{N_{l-1}} \mathbf{w}_l^{i,k} * \mathbf{f}_{l-1}^i)).
\end{equation}
Accordingly, an \textit{L}-layer CSEN network would have the following trainable weight and bias $\{\mathbf{w}, b\}$, parameters: $\mathbf{\Theta_{CSEN}}=\big\{ \{\mathbf{w}_1^i, b_1^i\}_{i=1}^{N_1}, \{\mathbf{w}_2^i, b_2^i\}_{i=1}^{N_2}, ... , \{\mathbf{w}_L^i, b_L^i\}_{i=1}^{N_L}\big\}$.

In SRC, the dictionary is collected by stacking training samples, for example, by concatenating the same class samples together. Thus, group $\ell_1$-minimization can be used instead of \eqref{lasso}:
\begin{equation}
    \min_\mathbf{x}  \left \{ \left \|  \mathbf{D}\mathbf{x}-\mathbf{y} \right \|_2^2 + \lambda \sum_{i=1}^{c}\left \|\mathbf{x}_{G,i} \right \|_2  \right \} 
\end{equation}
where $\mathbf{x}_{G,i}$ is the group of coefficients from class $i$. Therefore, the cost function for a CSEN can be expressed as,
\begin{equation}
\label{cost}
E(\mathbf{x}) =  \sum_p (\mathcal{P}_{\Theta}\left (\mathbf{\Tilde{x}} \right )_p- v_p)^2 + \lambda \sum_{i=1}^{c}\left \|\mathcal{P}_{\Theta}\left (\mathbf{\Tilde{x}} \right )_{G,i} \right \|_2. 
\end{equation}
where $\mathcal{P}_{\Theta}\left (\mathbf{\Tilde{x}} \right )_p$ and $v_p$ are the actual output and binary mask of the sparse code $\mathbf{x}$ for $p^{th}$ pixel, respectively.

The introduced regularization may bring additional computational complexity; hence, in the previous study \cite{csen}, an approximation of \eqref{cost} is adopted for CSEN by applying average pooling over the output and then performing SoftMax operation to produce the class probabilities directly. However, for the regression problem, which is undertaken in this study, we propose to modify the architecture by replacing the average pooling with the max pooling and inserting an additional convolutional layer and fully connected layer right after the max-pooling as illustrated in Fig. \ref{csen-diagram}. The included layers form the regression part of the modified CSEN. Then, the loss function of the modified CSEN for the regression can be expressed as $\mathcal{L}_{CSEN} = \sum_{i \in M} \textit{smooth}_{\ell_1}(\mathcal{P}_{\Theta}\left (\mathbf{\Tilde{x}}_i \right ) - d_i)$ over a batch $M$, where $\mathcal{P}_{\Theta}\left (\mathbf{\Tilde{x}}_i \right )$, $d_i$ are the predicted and real distance values for the $i^\text{th}$ object and smooth $\ell_1$-loss is expressed as,
\begin{subnumcases}
{\textit{smooth}_{\ell_1}(x) =}
   0.5x^2    \hfill & \text{ if $  \left | x \right | < 1$ } \\
   \left | x \right | - 0.5 & \text{ else }. 
\end{subnumcases}
Consequently, the proxy is selected as $\Tilde{\mathbf{x}}_i=\left ( \mathbf{D}^T \mathbf{D} + \lambda \mathbf{I} \right )^{-1} \mathbf{D}^T\mathbf{y}_i$ from LMMSE where $\mathbf{y}_i=\mathbf{A}\mathbf{f}_i$ is obtained for the extracted object feature $\mathbf{f}_i=\phi\left ( \mathbf{I}_i \right )$, and the input and output pair of the proposed method for the regression is $\left ( \Tilde{\mathbf{x}}^{train}, d^{train} \right )$ for the training.

\begin{figure}[]
  \includegraphics[width=1\linewidth]{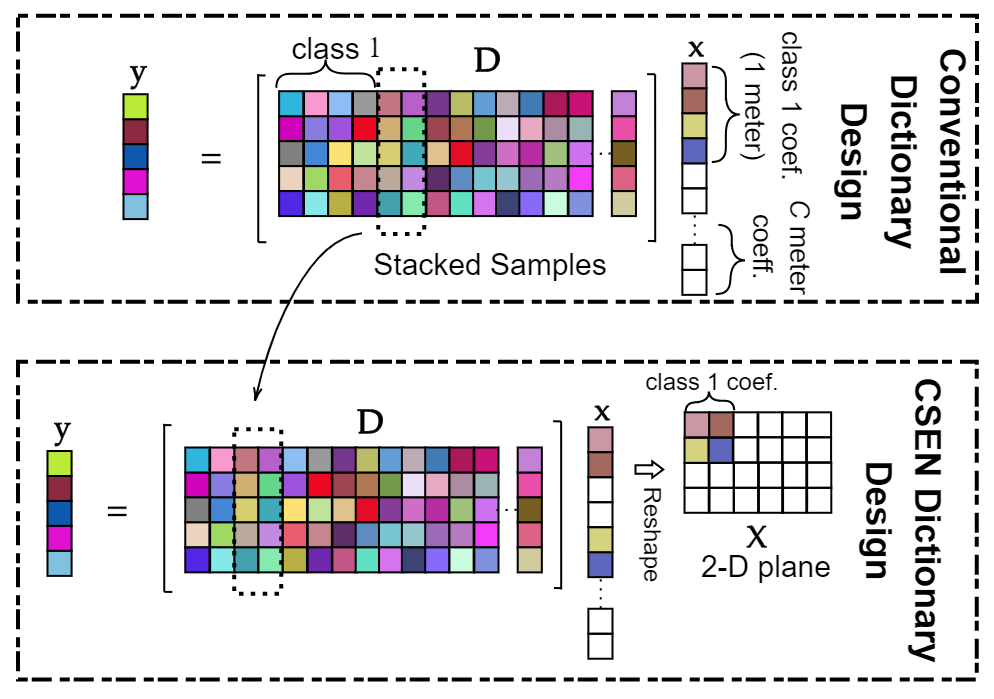}
  \caption{Conventional dictionary design versus the proposed dictionary design for the CSEN. In the conventional dictionary design, samples are collected with the increasing order of the distances. The first, second, third class categories correspond to 1m, 2m, 3m, respectively, and the $C^{\text{th}}$ class corresponds to $C$ meters.}
  \label{dictionary_design}
  \vspace{-0.5cm}
\end{figure}

Note the fact that the proposed RbR method can directly map the exact distance values and it is possible to train the model using the exact distance information. The quantized distances are only used when forming the dictionary $\mathbf{D}$ with the selected quantized dictionary samples. In Section \ref{rpc}, we have detailed the distance estimation utilizing representation-based classification with SRC and CRC approaches since they can only estimate the quantized distances that correspond to a classification task, i.e., class $c$ corresponds to objects of $c$ meter away from the camera. Therefore, the grouped features from different objects (e.g., car, person, and truck), but from the same distances ($c$-meter) as shown in Fig. \ref{dictionary_design_src_crc}. In this way, we will have a categorical invariant distance estimator unlike the literature work \cite{disnet,iccv}.

In the traditional approaches with SRC and CRC, one can directly use the collected representative dictionary $\mathbf{D}$ having the samples collected in a random order as long as the ordering is known since the recovery of $\mathbf{x}$ is obtained from $\mathbf{y = Dx}$. However, in the CSENs, direct mapping from $\mathbf{y}$ is performed using 2-D convolutional layers. Hence, in the proposed CSENs, it is important to group samples with the same quantized distances together after reshaping the proxy $\mathbf{\tilde{x}}$ since the grouped coefficients are max pooled in the feed-forward phase as discussed. Accordingly, the columns of the dictionary $\mathbf{D}$ are re-ordered in such a way that after reshaping the proxy into a 2-D plane, the samples with the same distances in the quantized level are grouped together. This proposed re-ordering topology is illustrated in Fig. \ref{dictionary_design} where 1-D coefficient vector $\mathbf{x}$ is reshaped to a 2-D plane that yields $\mathbf{X}$. Correspondingly, one can directly say that the input size of the CSEN depends on the collected dictionary size and the stride size (also kernel size) of the average pooling depends on the number of samples within the same distance level.

\subsection{Compressive Learning CSEN (CL-CSEN) Approach}

In the CSEN approach, the input is the reshaped proxy signal, $\mathbf{\tilde{x}}$, which is obtained directly by $\mathbf{\tilde{x}} = \left( \mathbf{D}^T\mathbf{D}+\lambda\mathbf{I} \right)^{-1}\mathbf{D}^T \mathbf{y}$. Ultimately, the performance of the CSEN was therefore limited to this proxy mapping stage i.e., $\mathbf{\Tilde{x}} = \mathbf{By}$ since $\mathbf{B}$ is treated as a constant during the training. To overcome this limitation, we propose to fine-tune the denoiser matrix $\mathbf{B}$ as follows: we include two additional fully-connected (dense) layers right before the first convolutional layer of the CSENs. The neurons connecting the input layer to the first hidden dense layer are initialized with $\mathbf{B}^T$ where $\mathbf{B}=\left( \mathbf{D}^T\mathbf{D}+\lambda\mathbf{I} \right)^{-1}\mathbf{D}^T$. Next, the output of the first hidden dense layer is reshaped to form the input of the first hidden convolutional layer.

\begin{figure}[h]
  \includegraphics[width=1\linewidth]{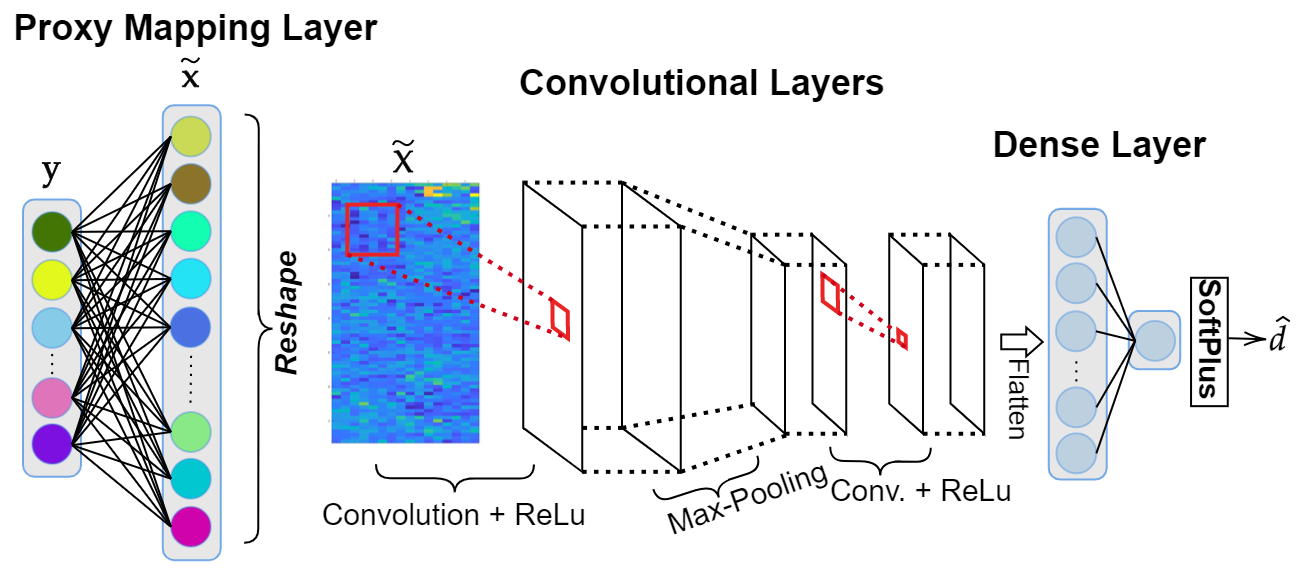}
  \caption{The proposed Compressive Learning CSEN (CL-CSEN) framework that jointly optimizes proxy mapping with support estimation and regression parts during the training.}
  \label{cl_csen}
\end{figure}

The CL-CSEN framework is illustrated in Fig. \ref{cl_csen} where the mapping from low-dimensional to high-dimensional space is learned during training. In this way, the proxy mapping layer is jointly optimized with the CSEN part of the CL-CSEN model to maximize the regression performance. Hence, the input and output pair of the proposed method with CL-CSEN will be $\left ( \mathbf{y}^{train}, d^{train} \right )$ for the training.

\section{Experimental Evaluation}
\label{results}

\begin{figure}[h]
\centering
  \includegraphics[width=0.85\linewidth]{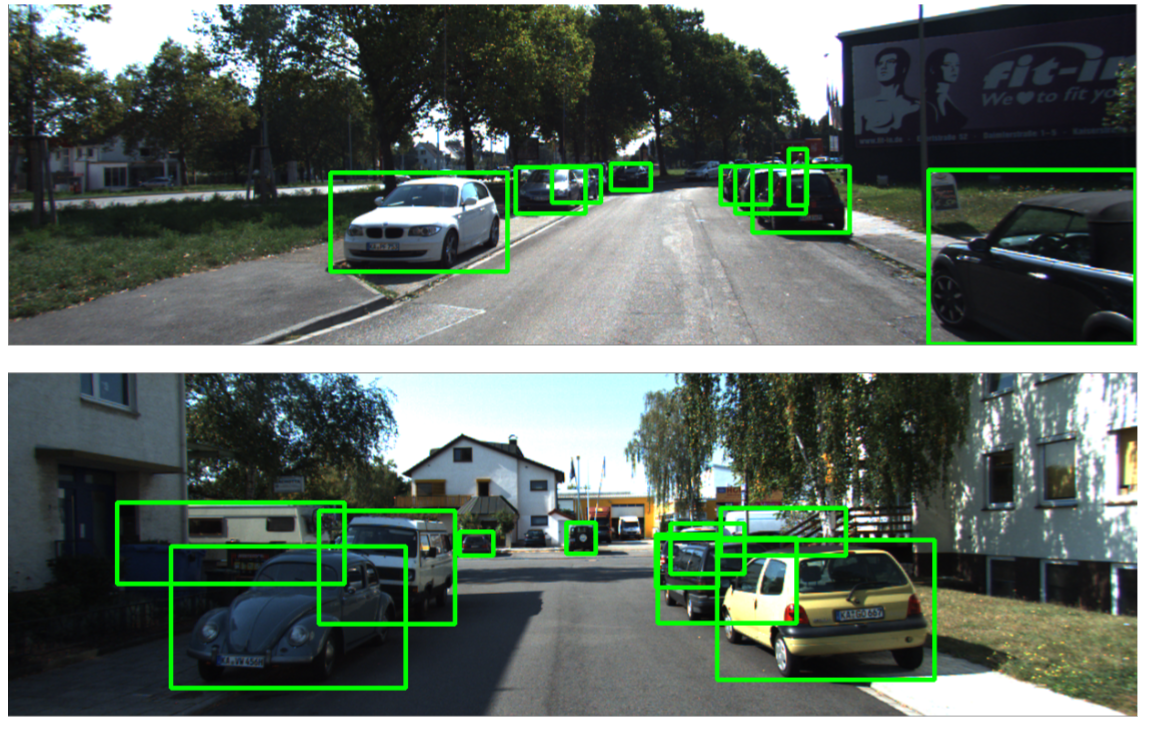}
  \caption{Two sample frames from KITTI 3D Object Detection showing overlapped samples.}
  \label{kitti_sample}
\end{figure}

The performance of the proposed approach is evaluated over the KITTI 3D Object Detection \cite{kitti} dataset. KITTI provides 3D bounding boxes for the detected objects as well as their categories. Besides having 3D object dimensions including length, height, and width, the dataset has the information of the 3D object locations: x,y, and z in camera coordinates. Hence, we use the z location information as the ground truth for the object distance estimation task. The collected frames are captured by a moving platform/vehicle from rural areas, a mid-size city, and highways. One challenge with this dataset is that there are overlapping samples on the observed scene as illustrated in Fig. \ref{kitti_sample}.

\subsection{Experimental Setup}

The KITTI annotations consist of 7481 images and there are a total of 40 570 objects having the distance information. The majority of them, 38 307 objects are in the range of $[0.5, 60.5]$ meters. In this study, the objects between the given range are selected for the evaluation in order to remove the outlier objects that are significantly far or close to the camera. The selected and cropped objects are then resized to $64\times64$ images and fed to the different feature extractor networks. We have created two different experimental setups. In the first one, a total of 19 769 samples are randomly selected for the training split and the remaining 18 538 samples are for the testing. In the second, only 4800 samples are used for training while the majority (33 507 samples) are used for the test. Consequently, these scenarios fulfill the aim of this study, i.e., evaluation of the learning capability with the limited amount of data (approximately 50\% and less than $13\%$ of the annotated data in the first and latter scenarios, respectively).

\subsubsection{CSEN and CL-CSEN Configurations}

To form the dictionary $\mathbf{D}$, we allocate 1200 samples from the training split and quantize those samples using 61 partitions in such a way that at the end, there are 20 samples per meter ($20 \times 60 = 1200$ objects in total) within the selected distance range. Recall the fact that these selected samples for each meter consist of different object categories such as person, car, truck, and trailer. Thus, $\mathbf{D}$ consists of 1200 samples in the proposed approaches with the CSEN and CL-CSEN. The compression ratio is set to $\text{CR}=m/d = 0.5$ using the PCA matrix $\mathbf{A}$ that is computed using the allocated samples for the dictionary formation. Consequently, the size of the equivalent dictionary $\mathbf{D}$ would be $m \times 1200$ with the proposed compression using PCA where $m=512$, $256$, and $1024$ for DenseNet-121, VGG19, and ResNet, respectively. Consequently, the corresponding denoiser matrix $\mathbf{B} = \left ( \mathbf{D}^T \mathbf{D} + \lambda \mathbf{I} \right )^{-1} \mathbf{D}^T$ would be $1200 \times m$. Therefore, the reshaped version of the computed proxy signal $\mathbf{\Tilde{x}} = \mathbf{By}$, $\mathbf{\Tilde{x}} \in \mathbb{R}^{n=1200}$ has the size of $80 \times 15$ in the 2-D plane. Finally, the remaining training samples are used for the training of CSEN and CL-CSEN.

The proposed compact CSEN structure given in Fig. \ref{csen-diagram} consists of only two convolutional (both with $5\times5$ filter sizes) and one dense layer. The first convolutional layer has 64 weight kernels that is followed by max-pooling with $4 \times 5$ pooling size. The second convolutional layer has only one kernel that creates a feature map that is flattened and connected to the single output neuron. In the CL-CSEN, there are additional two fully connected dense layers with the number of neurons corresponding to the size of $\mathbf{B}^T$ as previously discussed. In this way, the followed compact structure brings the ability to learn from a limited amount of data. All the layers have ReLu as the activation function except the output that has the SoftPlus activation function.

The CSEN is trained with 100 epochs and batch size of 16 by Adam optimizer \cite{kingma2014adam} using the proposed default parameter values as a learning rate $\alpha=10^{-3}$, $\beta_1 = 0.9$, and $\beta_2 = 0.999$. From the training set, we separate $20\%$ of samples for the validation to select the best network model to be used for testing. The experiments have been performed using Python on a PC with NVidia ® 1080 Ti GPU card, Intel ® i$9-7900$X CPU having 128 GB system memory. The CSEN and CL-CSEN are implemented with the Tensorflow library \cite{abadi2016tensorflow}. The hyper-parameter of $\lambda$ is first searched in log-scale within the range $\lambda^* \in [10^{-13}, 10^{3}]$. Afterwards, the fine-tuned version is set with few more steps by slight adjustment such that $\lambda = \lambda^* \pm 10^{log(\lambda^*)}$.

\subsubsection{Competing Methods}

\begin{table*}[ht]
\caption{The statistical (mean and standard deviations) performance metrics are reported from five different runs to show the object distance estimation performance of the proposed approach against the competing methods over the KITTI dataset and using different feature extractor networks, $\phi: \mathbb{R}^{N \times N \times 3} \rightarrow \mathbb{R}^{d}$. The train:test splits are selected as approximately 1:1 proportion. In the metrics, $\downarrow$:lower is better and $\uparrow$: higher is better.}
\centering
\label{table-results1}
\resizebox{\linewidth}{!}{
\begin{tabular}{@{}ccccccccc@{}}
\toprule
$\phi\left( . \right)$ & \textbf{Method} & ARD $\downarrow$ & SRD $\downarrow$ & RMSE $\downarrow$ & $\text{RMSE}_{{\textit{log}}}$ $\downarrow$ & $\delta < 1.25$ $\uparrow$ & $\delta < 1.25^2$ $\uparrow$ & $\delta < 1.25^3$ $\uparrow$ \\ 
\midrule
\multicolumn{1}{c}{\multirow{4}{*}{\rotatebox[origin=c]{90}{DenseNet-121}}} \vspace{0.1cm} & Support Vector Regressor (SVR) \cite{svr} & 0.2588 $\pm$ 0.003 & 1.7764 $\pm$ 0.041 & 5.3239 $\pm$ 0.013 & 0.4189 $\pm$ 0.006 & 0.6908 $\pm$ 0.002 & 0.8862 $\pm$ 0.003 & 0.9433 $\pm$ 0.003 \\

\vspace{0.1cm} & Base Model (CRC-light) \cite{collaborative} & 0.4183 $\pm$ 0.008 & 6.9585 $\pm$ 0.346 & 12.0007 $\pm$ 0.164 & 0.7462 $\pm$ 0.014 & 0.4447 $\pm$ 0.002 & 0.6821 $\pm$ 0.003 & 0.8055 $\pm$ 0.003 \\

\vspace{0.1cm} & \textbf{CSEN (Proposed)} & 0.2828 $\pm$ 0.006 & 2.2385 $\pm$ 0.091 & 6.2951 $\pm$ 0.061 & 0.4344 $\pm$ 0.052 & 0.6268 $\pm$ 0.008 & 0.8630 $\pm$ 0.003 & 0.9367 $\pm$ 0.002 \\ 

\vspace{0.1cm} & \textbf{CL-CSEN (Proposed)} & \textbf{0.2005} $\pm$ \textbf{0.009} & \textbf{1.2137} $\pm$ \textbf{0.084} & \textbf{4.3413} $\pm$ \textbf{0.048} & \textbf{0.2720} $\pm$ \textbf{0.014} & \textbf{0.7870} $\pm$ \textbf{0.006} & \textbf{0.9361} $\pm$ \textbf{0.005} & \textbf{0.9704} $\pm$ \textbf{0.003} \\ \midrule

\multicolumn{1}{c}{\multirow{5}{*}{\rotatebox[origin=c]{90}{VGG19}}}  \vspace{0.1cm} & Support Vector Regressor (SVR) \cite{svr} & 0.3496 $\pm$ 0.007 & 3.1122 $\pm$ 0.131 & 6.9459 $\pm$ 0.045 & 0.4690 $\pm$ 0.023 & 0.5752 $\pm$ 0.006 & 0.8325 $\pm$ 0.004 & 0.9172 $\pm$ 0.001 \\

\vspace{0.1cm} & Base Model (CRC-light) \cite{collaborative} & 0.4029 $\pm$ 0.003 & 6.1675 $\pm$ 0.134 & 12.2411 $\pm$ 0.112 & 0.8556 $\pm$ 0.031 & 0.4266 $\pm$ 0.004 & 0.6492 $\pm$ 0.006 & 0.7682 $\pm$ 0.007 \\

\vspace{0.1cm} & \textbf{CSEN (Proposed)} & 0.2917 $\pm$ 0.013 & 2.3542 $\pm$ 0.150 & 6.4498 $\pm$ 0.037 & 0.4581 $\pm$ 0.081 & 0.6058 $\pm$ 0.006 & 0.8510 $\pm$ 0.009 & 0.9307 $\pm$ 0.008 \\

\vspace{0.1cm} & \textbf{CL-CSEN (Proposed)} & \textbf{0.2221} $\pm$ \textbf{0.010} & \textbf{1.5034} $\pm$ \textbf{0.116} & \textbf{4.8132} $\pm$ \textbf{0.044} & \textbf{0.3021} $\pm$ \textbf{0.012} & \textbf{0.7448} $\pm$ \textbf{0.006} & \textbf{0.9164} $\pm$ \textbf{0.004} & \textbf{0.9623} $\pm$ \textbf{0.003} \\ \midrule

\multicolumn{1}{c}{\multirow{5}{*}{\rotatebox[origin=c]{90}{ResNet-50}}} \vspace{0.1cm} & Support Vector Regressor (SVR) \cite{svr} & 0.2509 $\pm$ 0.004 & 1.7669 $\pm$ 0.052 & 5.3531 $\pm$ 0.049 & 0.3613 $\pm$ 0.004 & 0.7004 $\pm$ 0.004 & 0.8989 $\pm$ 0.002 & 0.9519 $\pm$ 0.001 \\

\vspace{0.1cm} & Base Model (CRC-light) \cite{collaborative} & 0.3781 $\pm$ 0.004 & 5.6263 $\pm$ 0.146 & 10.9210 $\pm$ 0.082 & 0.6472 $\pm$ 0.011 & 0.4712 $\pm$ 0.002 & 0.7193 $\pm$ 0.001 & 0.8443 $\pm$ 0.002 \\

\vspace{0.1cm} & \textbf{CSEN (Proposed)} & 0.2400 $\pm$ 0.006 & 1.6777 $\pm$ 0.073 & 5.5212 $\pm$ 0.087 & 0.3459 $\pm$ 0.011 & 0.6902 $\pm$ 0.008 & 0.8983 $\pm$ 0.004 & 0.9533 $\pm$ 0.002 \\

\vspace{0.1cm} & \textbf{CL-CSEN (Proposed)} & \textbf{0.1934} $\pm$ \textbf{0.009} & \textbf{1.1710} $\pm$ \textbf{0.097} & \textbf{4.0849} $\pm$ \textbf{0.044} & \textbf{0.2604} $\pm$ \textbf{0.008} & \textbf{0.8148} $\pm$ \textbf{0.005} & \textbf{0.9439} $\pm$ \textbf{0.004} & \textbf{0.9730} $\pm$ \textbf{0.002} \\ \bottomrule
\end{tabular}
}
\end{table*}

Since we propose to use RbR by utilizing the regularized least-square sense solution as the coarse estimation of the support sets, the performance analysis will be performed against the base model with CRC \cite{collaborative}, and then, the improvement over CRC by the proposed CSEN and CL-CSEN will be reported. Moreover, we include various different solvers for SRC approach including ADMM \cite{ADMM}, Dalm \cite{fast}, OMP \cite{fast}, Homotopy \cite{homotopy}, GPSR \cite{gpsr}, L1LS \cite{l1ls}, $\ell_1$-magic \cite{l1magic}, Palm \cite{fast}. In addition, the performance evaluations are performed against the SVR \cite{svr} that has been used by \cite{iccv} and \cite{svr} for distance estimation. Note that compared to \cite{iccv} and \cite{svr}, we use the enhanced features obtained by the feature extraction method explained earlier. The SVR configuration is developed by searching the optimal hyper-parameters. Accordingly, the grid-search is applied over the validation set with the following kernel functions: linear, Radial Basis Function (RBF), and polynomial using the following parameters: $\gamma$ parameter (kernel coefficients for the RBF and polynomial kernels) in the range $[10^{-3}, 10^{3}]$ by varying in the log-scale, the degree of the polynomial \{$2$, $3$, $4$\}, the regularization parameter ($C$ parameter) in the range $[10^{-3}, 10^3]$ by varying in the log-scale.

To make a fair comparison with the competing methods, the training set of SVR includes also the dictionary samples in addition to the training samples that are used in the proposed CSEN and CL-CSEN. Similarly, the dictionary samples in SRC and CRC methods include the training samples plus the dictionary samples of CSEN and CL-CSEN. The same feature extraction procedure in the proposed method is used in the SRC, CRC, and SVR (i.e., $\phi\left ( \mathbf{I}_i \right )$ where $\phi$ is the pre-trained network for the cropped and resized object $\mathbf{I}_i$). The same CR is used by the PCA for SRC and CRC. In SVR, it is not feasible to compute the exact solution due to the scale of the data; and hence, we use Nystroem method \cite{nystrom1, nystrom2} for the kernel approximation in order to approximate $m=\text{CR} \times d$ number of feature maps where $\text{CR} = 0.5$. Overall, we keep the same CR value for all the methods in the experimental evaluations.

\subsection{Experimental Results}

The same performance metrics as used in \cite{plug-and-play,disnet,iccv,vid2depth,struct2depth,stereo} are used to evaluate the distance estimation performance of the proposed approach. Let the actual and predicted distances be $d_i$ and $\hat{d_i}$ and $N$ is the number of samples in the test split, then for a given threshold $t$, the metric Threshold is defined as,
\begin{equation}
    \text{\% of } \hat{d_i} ~s.t. ~\text{max}\left( \hat{d_i}/d_i, d_i/\hat{d_i} \right ) = \delta < t.
\end{equation}
Next, the Absolute Relative Distance (ARD) and the Squared Relative Distance (SRD) are defined as follows:
\begin{equation}
    \text{ARD} = \frac{1}{N}\sum_{i=1}^N \left( | \hat{d_i} - d_i | / d_i \right),
\end{equation}
\begin{equation}
    \text{SRD} = \frac{1}{N}\sum_{i=1}^N \left( (\hat{d_i} - d_i) ^2 / d_i \right).
\end{equation}
Finally, the Root of Mean Squared Error (RMSE) and the root of the Mean Squared logarithmic Error ($\text{RMSE}_\text{log}$) are,
\begin{equation}
    \text{RMSE} = \sqrt{\frac{1}{N}\sum_{i=1}^N \left( \hat{d_i} - d_i \right) ^2},
\end{equation}
\begin{equation}
    \text{RMSE}_\text{log} = \sqrt{\frac{1}{N}\sum_{i=1}^N \left( \log \hat{d_i} - \log d_i \right) ^2}.
\end{equation}

\begin{figure*}
     \centering
     \begin{subfigure}[b]{0.3\linewidth}
         \centering
         \includegraphics[width=\linewidth]{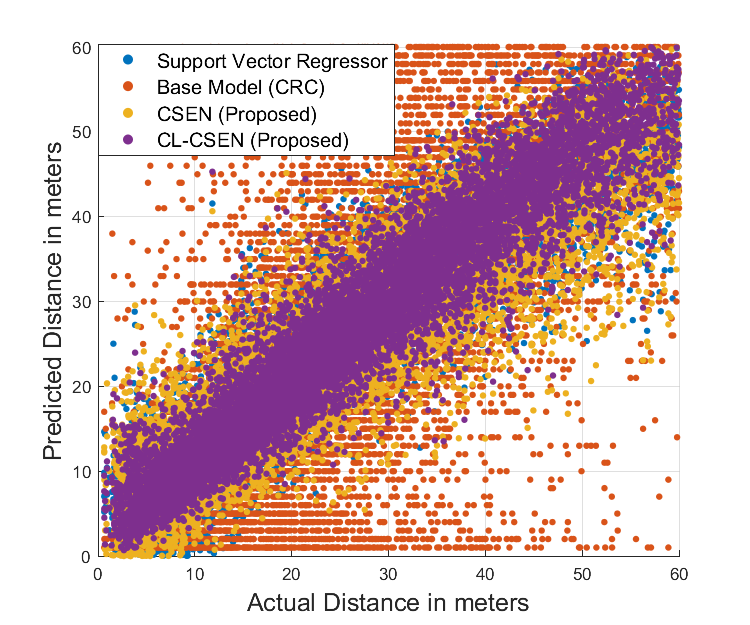}
         \caption{DenseNet-121 Feature Extraction}
         \label{fig:scattering-densenet121}
     \end{subfigure}
     \hfill
     \begin{subfigure}[b]{0.3\linewidth}
         \centering
         \includegraphics[width=\linewidth]{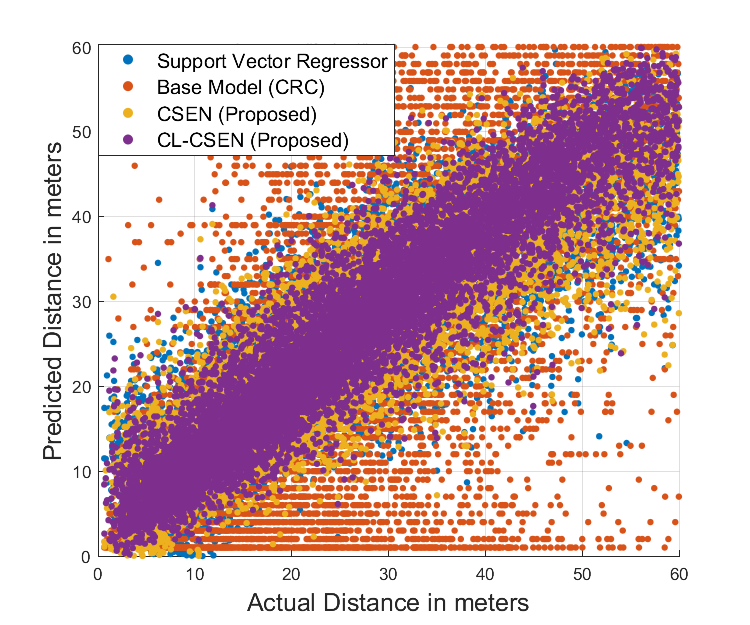}
         \caption{VGG-19 Feature Extraction.}
         \label{fig:scattering-vgg19}
     \end{subfigure}
     \hfill
     \begin{subfigure}[b]{0.3\linewidth}
         \centering
         \includegraphics[width=\linewidth]{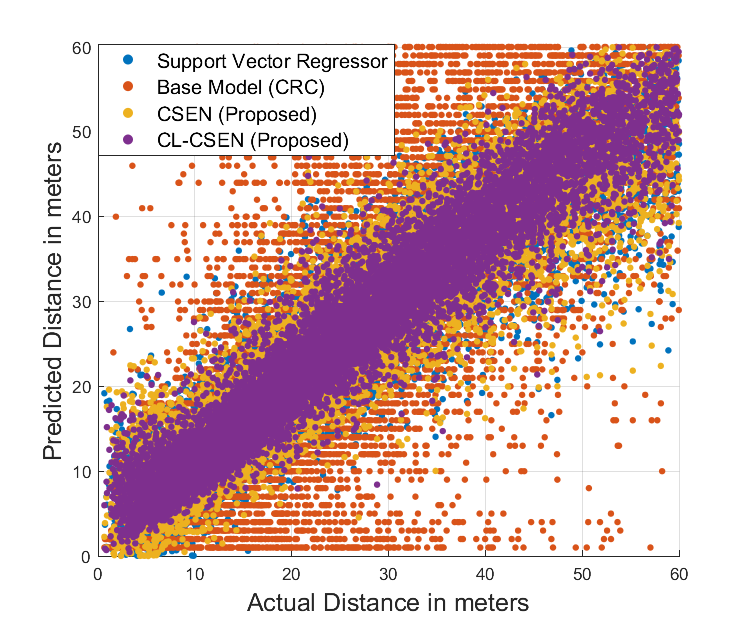}
         \caption{ResNet-50 Feature Extraction}
         \label{fig:scattering-resnet50}
     \end{subfigure}
        \caption{Predicted vs. actual distances of the objects in the test set for the proposed CSEN and CL-CSEN and compared methods using different feature extractor networks. In the scattering plot, each point represents a sample object in the KITTI dataset that is partitioned to train:test corresponding approximately 1:1 proportion.}
        \label{fig:scattering-plot-set1}
\end{figure*}

\begin{table*}[ht!]
\caption{The statistical (mean and standard deviations) performance metrics are reported from five different runs to show the object distance estimation performance of the proposed approach against the competing methods over the KITTI dataset and using different feature extractor networks, $\phi: \mathbb{R}^{N \times N \times 3} \rightarrow \mathbb{R}^{d}$. The train:test splits are selected as approximately 1:17 proportion and the selected distance sensitivity (with quantization) is 1m. In the metrics, $\downarrow$:lower is better and $\uparrow$: higher is better.}
\centering
\label{table-results2}
\resizebox{\linewidth}{!}{
\begin{tabular}{@{}ccccccccc@{}}
\toprule
$\phi\left( . \right)$ & \textbf{Method} & ARD $\downarrow$ & SRD $\downarrow$ & RMSE $\downarrow$ & $\text{RMSE}_{{\textit{log}}}$ $\downarrow$ & $\delta < 1.25$ $\uparrow$ & $\delta < 1.25^2$ $\uparrow$ & $\delta < 1.25^3$ $\uparrow$ \\ 
\midrule
\multicolumn{1}{c}{\multirow{18}{*}{\rotatebox[origin=c]{90}{DenseNet-121}}} \vspace{0.2cm} & CRC-light \cite{collaborative} & 0.4157 $\pm$ 0.0.10 & 6.9163 $\pm$ 0.362 & 12.0034 $\pm$ 0.159 & 0.7442 $\pm$ 0.010 & 0.4337 $\pm$ 0.002 & 0.6802 $\pm$ 0.004 & 0.8018 $\pm$ 0.003 \\

\vspace{0.2cm} & CRC \cite{collaborative} & 0.3384 $\pm$ 0.003 & 4.9194 $\pm$ 0.110 & 11.1735 $\pm$ 0.068 & 0.9687 $\pm$ 0.007 & 0.5060 $\pm$ 0.002 & 0.7091 $\pm$ 0.003 & 0.7887 $\pm$ 0.002 \\

\vspace{0.2cm} & ADMM \cite{ADMM} & 0.3662 $\pm$ 0.001 & 5.7964 $\pm$ 0.109 & 9.7512 $\pm$ 0.092 & 0.5286 $\pm$ 0.004 & 0.5429 $\pm$ 0.002 & 0.7731 $\pm$ 0.002 & 0.8737 $\pm$ 0.001 \\

\vspace{0.2cm} & Dalm \cite{fast} & 0.3502 $\pm$ 0.003 & 5.3903 $\pm$ 0.108 & 9.7558 $\pm$ 0.107 & 0.5566 $\pm$ 0.006 & 0.5498 $\pm$ 0.003 & 0.7753 $\pm$ 0.003 & 0.8706 $\pm$ 0.002 \\

\vspace{0.2cm} & OMP \cite{fast} & 0.4279 $\pm$ 0.004 & 8.0630 $\pm$ 0.142 & 11.1767 $\pm$ 0.072 & 0.6326 $\pm$ 0.005 & 0.5203 $\pm$ 0.003 & 0.7386 $\pm$ 0.003 & 0.9019 $\pm$ 0.002 \\

\vspace{0.2cm} & Homotopy \cite{homotopy} & 0.3747 $\pm$ 0.004 & 5.8120 $\pm$ 0.075 & 9.5982 $\pm$ 0.053 & 0.4917 $\pm$ 0.003 & 0.5415 $\pm$ 0.003 & 0.7764 $\pm$ 0.003 & 0.8806 $\pm$ 0.002 \\

\vspace{0.2cm} & GPSR \cite{gpsr} & 0.3357 $\pm$ 0.003 & 4.9460 $\pm$ 0.099 & 9.4322 $\pm$ 0.089 & 0.5456 $\pm$ 0.003 & 0.5547 $\pm$ 0.001 & 0.7824 $\pm$ 0.001 & 0.8777 $\pm$ 0.001 \\

\vspace{0.2cm} & L1LS \cite{l1ls} & 0.3550 $\pm$ 0.003 & 5.4403 $\pm$ 0.096 & 11.9869 $\pm$ 0.090 & 1.0916 $\pm$ 0.010 & 0.4932 $\pm$ 0.002 & 0.6840 $\pm$ 0.004 & 0.7584 $\pm$ 0.004 \\

\vspace{0.2cm} & $\ell_1$-magic \cite{l1magic} & 0.3579 $\pm$ 0.004 & 5.6748 $\pm$ 0.141 & 9.9567 $\pm$ 0.114 & 0.5695 $\pm$ 0.006 & 0.5457 $\pm$ 0.003 & 0.7695 $\pm$ 0.002 & 0.8655 $\pm$ 0.001 \\

\vspace{0.2cm} & Palm \cite{fast} & 0.3262 $\pm$ 0.003 & 4.6281 $\pm$ 0.109 & 10.6396 $\pm$ 0.075 & 0.8858 $\pm$ 0.006 & 0.5278 $\pm$ 0.003 & 0.6919 $\pm$ 0.082 & 0.8117 $\pm$ 0.002 \\

\vspace{0.2cm} & \textbf{CSEN (Proposed)} & 0.3308 $\pm$ 0.036 & 3.0487 $\pm$ 0.534 & 6.9950 $\pm$ 0.367 & 0.5370 $\pm$ 0.217 & 0.5668 $\pm$ 0.023 & 0.8215 $\pm$ 0.025 & 0.9132 $\pm$ 0.020 \\

\vspace{0.2cm} & \textbf{CL-CSEN (Proposed)} & \textbf{0.3167} $\pm$ \textbf{0.006} & \textbf{2.8268} $\pm$ \textbf{0.083} & \textbf{6.2036} $\pm$ \textbf{0.046} & \textbf{0.3753} $\pm$ \textbf{0.017} & \textbf{0.6378} $\pm$ \textbf{0.004} & \textbf{0.8555} $\pm$ \textbf{0.004} & \textbf{0.9313} $\pm$ \textbf{0.002} \\ \midrule

\multicolumn{1}{c}{\multirow{19}{*}{\rotatebox[origin=c]{90}{VGG19}}}  \vspace{0.2cm} & CRC-light \cite{collaborative} & 0.4018 $\pm$ 0.002 & 6.1850 $\pm$ 0.083 & 12.2543 $\pm$ 0.101 & 0.8549 $\pm$ 0.032 & 0.4163 $\pm$ 0.004 & 0.6498 $\pm$ 0.005 & 0.7664 $\pm$ 0.007 \\

\vspace{0.2cm} & CRC \cite{collaborative} & 0.3591 $\pm$ 0.004 & 5.3996 $\pm$ 0.057 & 12.1625 $\pm$ 0.075 & 1.0796 $\pm$ 0.016 & 0.4727 $\pm$ 0.002 & 0.6576 $\pm$ 0.006 & 0.7333 $\pm$ 0.008 \\

\vspace{0.2cm} & ADMM \cite{ADMM} & 0.3506 $\pm$ 0.005 & 5.3399 $\pm$ 0.138 & 9.4499 $\pm$ 0.066 & 0.5114 $\pm$ 0.005 & 0.5547 $\pm$ 0.001 & 0.7829 $\pm$ 0.003 & 0.8799 $\pm$ 0.001 \\

\vspace{0.2cm} & Dalm \cite{fast} & 0.3535 $\pm$ 0.004 & 5.4561 $\pm$ 0.129 & 9.8062 $\pm$ 0.062 & 0.5653 $\pm$ 0.006 & 0.5466 $\pm$ 0.002 & 0.7697 $\pm$ 0.002 & 0.8656 $\pm$ 0.002 \\

\vspace{0.2cm} & OMP \cite{fast} & 0.3946 $\pm$ 0.004 & 6.7427 $\pm$ 0.119 & 10.2869 $\pm$ 0.028 & 0.5589 $\pm$ 0.004 & 0.5395 $\pm$ 0.003 & 0.7621 $\pm$ 0.002 & 0.8598 $\pm$ 0.001 \\

\vspace{0.2cm} & Homotopy \cite{homotopy} & 0.3532 $\pm$ 0.005 & 5.2429 $\pm$ 0.186 & 9.1591 $\pm$ 0.039 & 0.4624 $\pm$ 0.002 & 0.5604 $\pm$ 0.002 & 0.7931 $\pm$ 0.002 & 0.8921 $\pm$ 0.001 \\

\vspace{0.2cm} & GPSR \cite{gpsr} & 0.3301 $\pm$ 0.003 & 4.7034 $\pm$ 0.108 & 9.3988 $\pm$ 0.062 & 0.5679 $\pm$ 0.007 & 0.5540 $\pm$ 0.002 & 0.7773 $\pm$ 0.002 & 0.8711 $\pm$ 0.001 \\

\vspace{0.2cm} & L1LS \cite{l1ls} & 0.3683 $\pm$ 0.005 & 5.6823 $\pm$ 0.077 & 12.5880 $\pm$ 0.092 & 1.1247 $\pm$ 0.018 & 0.4641 $\pm$ 0.003 & 0.6432 $\pm$ 0.007 & 0.7166 $\pm$ 0.009 \\

\vspace{0.2cm} & $\ell_1$-magic \cite{l1magic} & 0.3541 $\pm$ 0.004 & 5.4787 $\pm$ 0.118 & 9.8251 $\pm$ 0.058 & 0.5669 $\pm$ 0.006 & 0.5464 $\pm$ 0.002 & 0.7692 $\pm$ 0.002 & 0.8652 $\pm$ 0.001 \\

\vspace{0.2cm} & Palm \cite{fast} & 0.3175 $\pm$ 0.001 & 4.2531 $\pm$ 0.026 & 10.3851 $\pm$ 0.041 & 0.8573 $\pm$ 0.009 & 0.5261 $\pm$ 0.001 & 0.7267 $\pm$ 0.003 & 0.8059 $\pm$ 0.005 \\

\vspace{0.2cm} & \textbf{CSEN (Proposed)} & 0.3401 $\pm$ 0.039 & 3.1667 $\pm$ 0.563 & 7.2027 $\pm$ 0.331 & 0.6763 $\pm$ 0.264 & 0.5392 $\pm$ 0.019 & 0.7978 $\pm$ 0.021 & 0.9000 $\pm$ 0.017 \\ 

\vspace{0.2cm} & \textbf{CL-CSEN (Proposed)} & \textbf{0.3062} $\pm$ \textbf{0.010} & \textbf{2.6452} $\pm$ \textbf{0.140} & \textbf{6.3759} $\pm$ \textbf{0.122} & \textbf{0.4222} $\pm$ \textbf{0.059} & \textbf{0.6091} $\pm$ \textbf{0.009} & \textbf{0.8404} $\pm$ \textbf{0.009} & \textbf{0.9265} $\pm$ \textbf{0.006} \\ \midrule

\multicolumn{1}{c}{\multirow{19}{*}{\rotatebox[origin=c]{90}{ResNet-50}}} \vspace{0.2cm} & CRC-light \cite{collaborative} & 0.3752 $\pm$ 0.003 & 5.5853 $\pm$ 0.081 & 10.8963 $\pm$ 0.066 & 0.6454 $\pm$ 0.014 & 0.4605 $\pm$ 0.001 & 0.7184 $\pm$ 0.003 & 0.8410 $\pm$ 0.002 \\

\vspace{0.2cm} & CRC \cite{collaborative} & 0.2817 $\pm$ 0.002 & 3.3945 $\pm$ 0.063 & 9.1777 $\pm$ 0.080 & 0.7371 $\pm$ 0.013 & 0.5598 $\pm$ 0.003 & 0.7786 $\pm$ 0.005 & 0.8562 $\pm$ 0.004 \\

\vspace{0.2cm} & ADMM \cite{ADMM} & 0.3155 $\pm$ 0.003 & 4.2173 $\pm$ 0.062 & 8.6938$\pm$ 0.062 & 0.4798 $\pm$ 0.006 & 0.5680 $\pm$ 0.002 & 0.8038 $\pm$ 0.002 & 0.8979 $\pm$ 0.002 \\

\vspace{0.2cm} & Dalm \cite{fast} & 0.2916 $\pm$ 0.003 & 3.6398 $\pm$ 0.075 & 8.4626 $\pm$ 0.082 & 0.4981 $\pm$ 0.008 & 0.5791 $\pm$ 0.003 & 0.8128 $\pm$ 0.004 & 0.9019 $\pm$ 0.002 \\

\vspace{0.2cm} & OMP \cite{fast} & 0.3352 $\pm$ 0.003 & 4.9965 $\pm$ 0.089 & 9.5783 $\pm$ 0.067 & 0.5639 $\pm$ 0.005 & 0.5550 $\pm$ 0.003 & 0.7813 $\pm$ 0.004 & 0.8761 $\pm$ 0.002 \\

\vspace{0.2cm} & Homotopy \cite{homotopy} & 0.3239 $\pm$ 0.005 & 4.2413 $\pm$ 0.103 & 8.4395 $\pm$ 0.026 & 0.4424 $\pm$ 0.004 & 0.5711 $\pm$ 0.001 & 0.8098 $\pm$ 0.002 & 0.9069 $\pm$ 0.002 \\

\vspace{0.2cm} & GPSR \cite{gpsr} & 0.2928 $\pm$ 0.003 & 3.6532 $\pm$ 0.094 & 8.4384 $\pm$ 0.100 & 0.4963 $\pm$ 0.007 & 0.5791 $\pm$ 0.003 & 0.8127 $\pm$ 0.003 & 0.9012 $\pm$ 0.002 \\

\vspace{0.2cm} & L1LS \cite{l1ls} & 0.2849 $\pm$ 0.004 & 3.4606 $\pm$ 0.090 & 9.4654 $\pm$ 0.078 & 0.7684 $\pm$ 0.011 & 0.5540 $\pm$ 0.003 & 0.7657 $\pm$ 0.005 & 0.8419 $\pm$ 0.004 \\

\vspace{0.2cm} & $\ell_1$-magic \cite{l1magic} & 0.2940 $\pm$ 0.002 & 3.6942 $\pm$ 0.057 & 8.5528 $\pm$ 0.059 & 0.5061 $\pm$ 0.007 & 0.5758 $\pm$ 0.002 & 0.8083 $\pm$ 0.003 & 0.8979 $\pm$ 0.002 \\

\vspace{0.2cm} & Palm \cite{fast} & 0.2767 $\pm$ 0.003 & 3.2185 $\pm$ 0.090 & 8.9170 $\pm$ 0.088 & 0.6784 $\pm$ 0.012 & 0.5668 $\pm$ 0.003 & 0.7855 $\pm$ 0.004 & 0.8643 $\pm$ 0.004 \\

\vspace{0.2cm} & \textbf{CSEN (Proposed)} & 0.2835 $\pm$ 0.035 & 2.2479 $\pm$ 0.437 & 6.2142 $\pm$ 0.333 & 0.5074 $\pm$ 0.155 & 0.6076 $\pm$ 0.020 & 0.8481 $\pm$ 0.016 & 0.9280 $\pm$ 0.011 \\

\vspace{0.2cm} & \textbf{CL-CSEN (Proposed)} & \textbf{0.3359} $\pm$ \textbf{0.010} & \textbf{2.9720} $\pm$ \textbf{0.106} & \textbf{6.0765} $\pm$ \textbf{0.041} & \textbf{0.3735} $\pm$ \textbf{0.005} & \textbf{0.6398} $\pm$ \textbf{0.005} & \textbf{0.8510} $\pm$ \textbf{0.004} & \textbf{0.9261} $\pm$ \textbf{0.003} \\ \bottomrule
\end{tabular}
}
\end{table*}

\begin{figure*}[h!]
     \centering
     \begin{subfigure}[b]{0.3\linewidth}
         \centering
         \includegraphics[width=\linewidth]{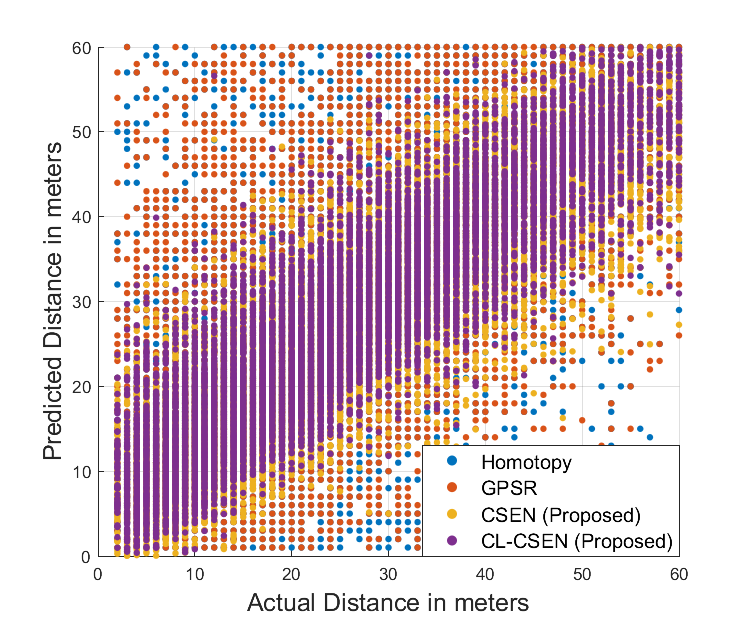}
         \caption{DenseNet-121 Feature Extraction}
         \label{fig:scattering-ablation-densenet121}
     \end{subfigure}
     \hfill
     \begin{subfigure}[b]{0.3\linewidth}
         \centering
         \includegraphics[width=\linewidth]{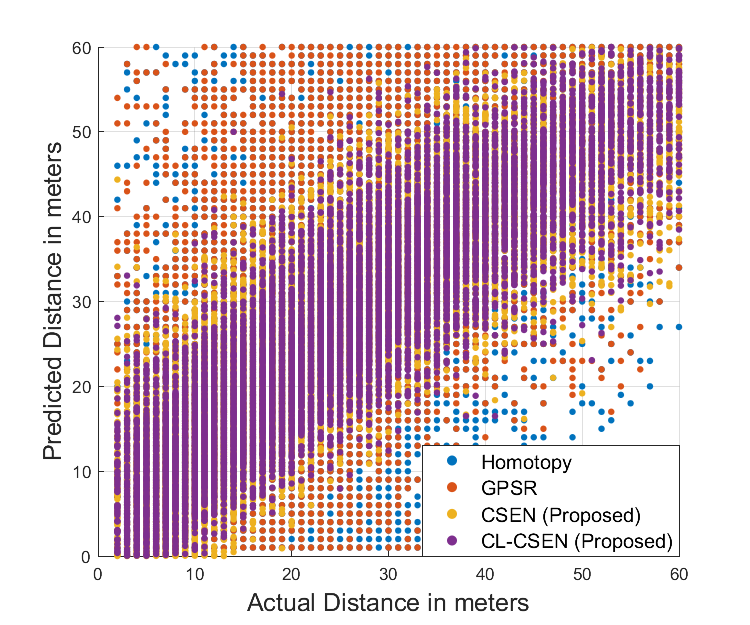}
         \caption{VGG-19 Feature Extraction.}
         \label{fig:scattering-ablation-vgg19}
     \end{subfigure}
     \hfill
     \begin{subfigure}[b]{0.3\linewidth}
         \centering
         \includegraphics[width=\linewidth]{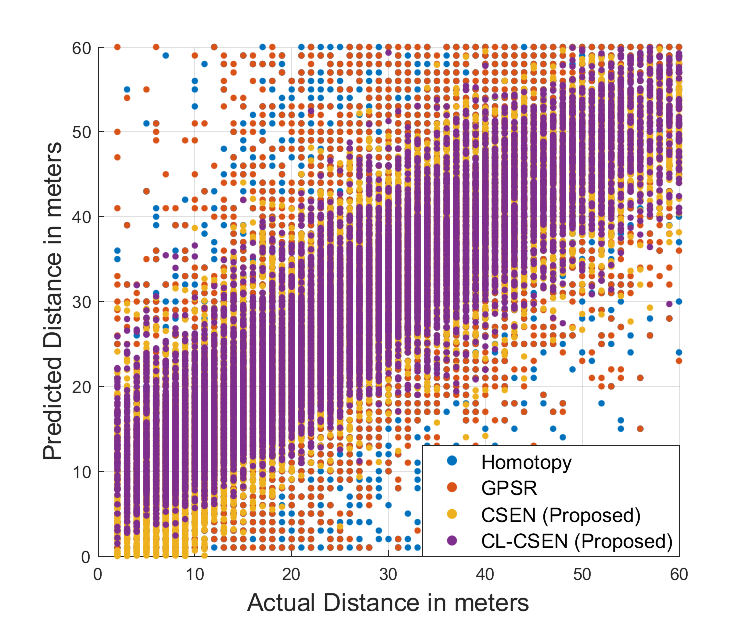}
         \caption{ResNet-50 Feature Extraction}
         \label{fig:scattering-ablation-resnet50}
     \end{subfigure}
        \caption{Predicted vs. actual distances of the objects in the test set for the proposed CSEN and CL-CSEN and compared methods using different feature extractor networks. In the scattering plot, each point represents a sample object in the KITTI dataset that is partitioned to train:test corresponding approximately 1:17 proportion. The selected distance sensitivity (with quantization) is one meter.}
        \label{fig:scattering-plot-set2}
\end{figure*}

The distance estimation performance of the proposed method is presented in Table \ref{table-results1}. In these results, we report the RbR performance with the proposed CSEN and CL-CSEN models where the quantization is not applied for the training and testing samples, but is only used for the dictionary reconstruction. Correspondingly, the train:test splits are chosen as approximately 1:1 proportion. Note the fact that multiple pre-trained networks are utilized for feature extraction. In this way, we aim to evaluate the performance effect of different network architectures in feature extraction. DenseNet-121 has skip-connections that connect each layer to every other layer so that each layer is densely connected, ResNet-50 only has skip-connections between every second layer, and VGG-19 does not have any such shortcut connection between the layers. Based on Table \ref{table-results1}, a higher estimation accuracy is achieved by the proposed approach compared to SVR and the performance is highly improved compared to our base CRC model. Moreover, the proposed method outperforms \cite{iccv} even though they use additional information such as the categorical class information of the objects and the  projection matrix for the training. For a more fair comparison, the proposed method is also compared with the base model of \cite{iccv} without classification; and the performance gap becomes even higher as expected. Additionally, scattering plots are provided in Fig. \ref{fig:scattering-plot-set1} demonstrating the actual distance versus the predicted distance by all methods. Correspondingly, we expect to see an identity transformation ideally. In the plots, the sample point sizes are purposely selected bigger to better illustrate the misdetections. Thus, considering the number of samples, most of them located at the identity line region and give a constant color view. Hence, it is observed that the CL-CSEN method provides the least scattered samples compared to the other methods especially when ResNet-50 features are used. Even though the reported metrics in Table \ref{table-results1} show improvements achieved by the proposed method; the performance gain is more visible for distant objects considering that the gap is significant in the squared metrics.

\begin{figure*}[ht]
  \includegraphics[width=1\linewidth]{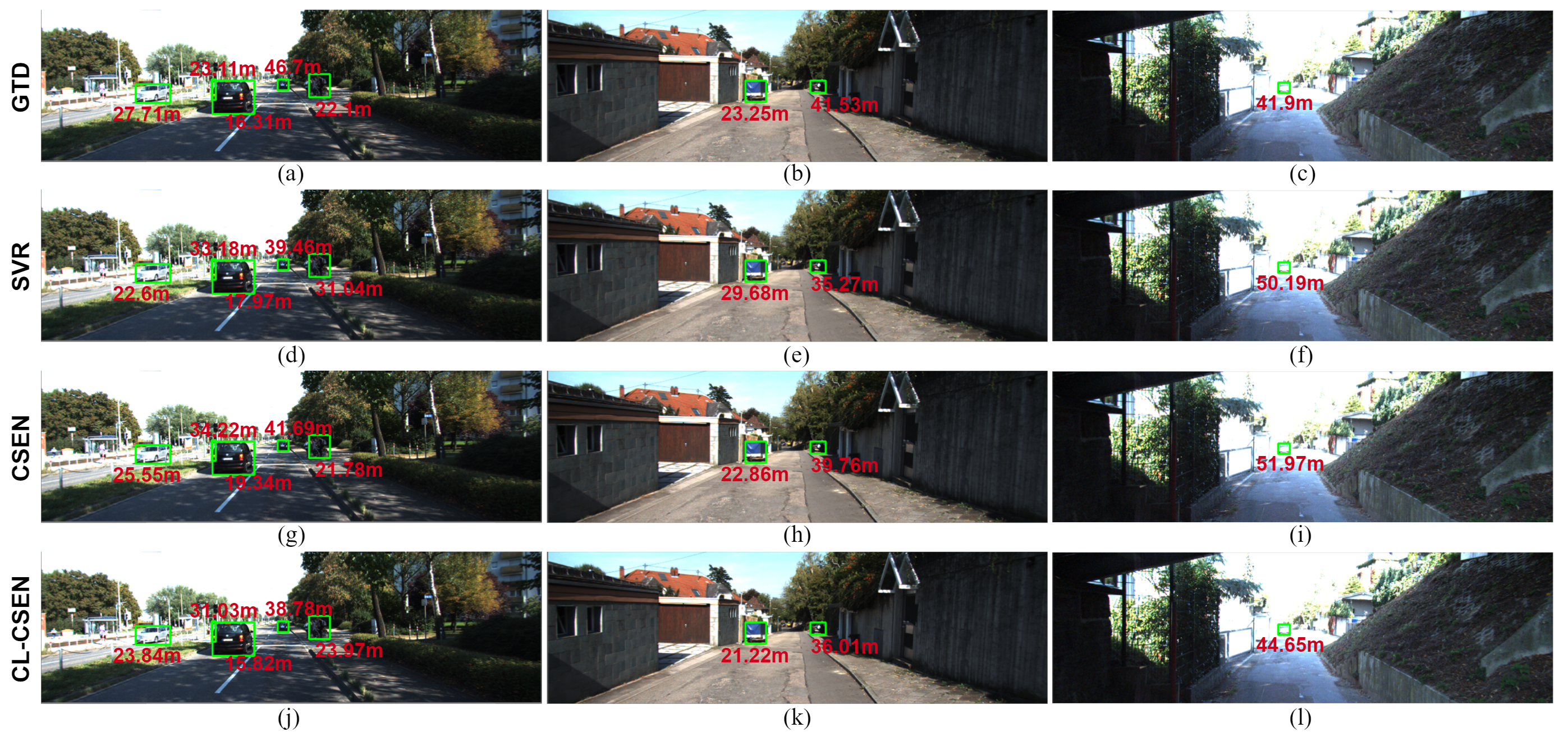}
  \caption{Three sample frames are shown with the object bounding boxes and their corresponding ground-truth distances (GTD) in the first row: (a), (b), and (c). Then, the estimated distances for the objects by the three best-performing methods in this work: SVR, CSEN, and CL-CSEN are illustrated in the second (d - e - f), third (g - h - i), and the last (j - k - l) rows, respectively. The approximate 1:1 ratio is followed in train:test splits.}
  \vspace{-0.25cm}
  \label{samples}
\end{figure*}

Next, the performance comparison is provided in Table \ref{table-results2} regarding the proposed method with CSEN and CL-CSEN versus CRC and SRC approaches. In this set of experiments, contrary to Table \ref{table-results1}, we have applied quantization to the training and testing samples of the CSEN and CL-CSEN approaches as previously discussed. Even though they can be trained for the full regression task, the competing methods including SRC and CRC do not have this ability; and hence, we wanted to compare the proposed approach fairly with them. In the table, CRC-light corresponds to our coarse estimation for the CSEN and CL-CSEN methods where the same number of dictionary samples are used in the CSEN approach. For the other SRC and CRC approaches, a total of 4800 samples are used to build the dictionary which corresponds to 1200 dictionary + 3600 training samples in CSEN and CL-CSEN. Based on Table \ref{table-results2}, it is clear that both proposed methods have achieved a significant performance gap over the competing methods. It is also shown that the CSEN and CL-CSEN methods are able to learn from such a limited number of training samples (only 4800 samples are used for training compared to 33 507 testing samples). Similarly, the scattering plot is provided in Fig. \ref{fig:scattering-plot-set2} for the second set of the experiments and the best-performing methods from Table \ref{table-results2}. Based on the plots, CL-CSEN has less distributed scatters due to the improved distance estimation performance: most of the test samples are overlapped near the identity transformation where few test samples are separated or distinguishable from the others since they are detected in error and located far from the overlapped points. The visual difference from the previous plot in Fig. \ref{fig:scattering-plot-set1} is that expectedly, the samples are located with 1m distances due to the applied quantization.

Three sample frames are shown in Fig. \ref{samples} with their corresponding true and the estimated object-specific distances by the three best-performing methods: SVR, CSEN, and CL-CSEN. The first frame (first column in Fig. \ref{samples}) represents a typical sample from the KITTI 3D Object Detection dataset in which there are overlapping objects. Moreover, even though the dataset may have some well-separated samples, the illumination conditions make it harder to perform analysis as observed in the third sample frame in Fig. \ref{samples}. Visual inspection based on these frames indicates that even though CL-CSEN provides enhanced performance than CSEN according to the quantitative analysis, for the close objects, CSEN seems to provide more accurate results. However, CSEN starts to underperform compared to CL-CSEN when the objects are distant from the camera.

\subsection{Computational Complexity Analysis}

\begin{table}[h!]
\centering
\caption{The number of trainable parameters is given in (a) for the proposed CSEN and CL-CSEN models. The elapsed times using the aforementioned PC setup are given in (b) for the methods.}
\small
\begin{subtable}{.48\textwidth}
\caption{The trainable parameters using different feature extractors $\phi$.}
\centering
 \resizebox{0.95\textwidth}{!}{
\begin{tabular}{|c|c|c|}
\hline
\rowcolor[gray]{.85} $\phi: \mathbb{R}^{N \times N \times 3} \rightarrow \mathbb{R}^{d}$ & Model & Number of Parameters \\ \hline \hline
$d \in \left \{1024,512,2048 \right \}$ & CSEN & 3,326 \\ \hline
\rowcolor[gray]{.98} $d=1024$ & CL-CSEN & 618,926 \\ \hline
$d=512$ & CL-CSEN & 311,726 \\ \hline
\rowcolor[gray]{.98} $d=2048$ & CL-CSEN & 1,233,326 \\ \hline
\end{tabular}}
\label{paramaters}
\end{subtable}

\bigskip
\noindent
\begin{subtable}{.48\textwidth}
\caption{Average elapsed times in milliseconds (ms) for the estimation of a test object sample. The given computational times are obtained in the case of ResNet-50 features.}
\centering
\resizebox{0.45\textwidth}{!}{
\begin{tabular}{|c|c|}
\hline
\rowcolor[gray]{.85} Method & Time (ms) \\ \hline \hline
SVR & 0.0035 \\ \hline
CRC-light & 2.0242 \\ \hline
\rowcolor[gray]{.98} CRC & 14.258 \\ \hline
ADMM & 198.14 \\ \hline
\rowcolor[gray]{.98} Dalm & 3574.0 \\ \hline
OMP & 241.22 \\ \hline
\rowcolor[gray]{.98}Homotopy & 30.591 \\ \hline
GPSR & 1547.0 \\ \hline
\rowcolor[gray]{.98} L1LS & 223.84 \\ \hline
$\ell_1$-magic & 2698.2 \\ \hline
\rowcolor[gray]{.98} Palm & 10996.0 \\ \hline
CSEN & 0.0348 \\ \hline
\rowcolor[gray]{.98} CL-CSEN & 0.0320 \\ \hline
\end{tabular}}
\label{telapsed}
\end{subtable}
\label{time_complexity}
\end{table}

The number of trainable parameters is provided in Table \ref{paramaters} for the proposed CSEN and CL-CSEN models. Accordingly, the CSEN model has only a few thousand trainable parameters since the denoiser matrix $\mathbf{B}$ is not trainable, whereas in the CL-CSEN model, depending on the size of $\mathbf{B}$ the trainable parameters vary. Nevertheless, both are still compact architectures only with a few layers. The elapsed times are reported in Table \ref{telapsed} on the aforementioned PC setup. On the other hand, SRC methods suffer drastic time complexity whereas elapsed times for CSEN and CL-CSEN methods are comparable with the SVR method. Note the fact that even though the CL-CSEN pipeline has more trainable parameters, the required time for the inference is less than CSEN. Because; the proxy mapping and reshaping stages for the following convolutional layers are implemented on GPU as an end-to-end pipeline that brings the computational efficiency that was lacking in the initial CSEN approach. Note that even though CSEN and CL-CSEN utilize the proxy mapping stage of the CRC approach, they are still computationally efficient because CRC-light and CRC require additional residual finding step that was explained in Section \ref{src-crc-approaches}.

\subsection{Discussion: 1-D versus 2-D Proxy Signal Representation}

The presented distance estimation results are obtained using the proposed CSEN and CL-CSEN approaches that contain 2-D convolution operations. One can investigate operating directly over the 1-D proxy signal without further reshaping it as the input of the first convolutional layer. Hence, we present the distance estimation results in Table \ref{1D-CSEN} using 1-D convolutional layers in the proposed approaches. Accordingly, the CSEN-1D and CL-CSEN-1D models do not have any reshaping operations contrary to 2-D versions illustrated in Fig. \ref{csen-diagram} and Fig. \ref{cl_csen}. It is observed that using the same number of trainable parameters, i.e., $25 \times 1$ filter sizes for each convolutional layer, the comparable results are obtained by performing 1-D inference on the proxy signal.

\begin{table*}[]
\centering
\caption{The statistical (mean and standard deviations) performance metrics are reported from five different runs using the 1D versions of the proposed approaches (CSEN-1D and CL-CSEN-1D) over the KITTI dataset and using different feature extractor networks, $\phi: \mathbb{R}^{N \times N \times 3} \rightarrow \mathbb{R}^{d}$. In the metrics, $\downarrow$:lower is better and $\uparrow$: higher is better.}
\begin{subtable}{1\textwidth}
\centering
\caption{The train:test splits are selected as approximately 1:1 proportion.}
\resizebox{.8\linewidth}{!}{
\begin{tabular}{c|cc|cc|cc}
        $\phi(.)$ & \multicolumn{2}{c|}{DenseNet-121} & \multicolumn{2}{c|}{VGG19} & \multicolumn{2}{c}{ResNet-50} \\ \hline
        & \textbf{CSEN-1D}       & \textbf{CL-CSEN-1D}       & \textbf{CSEN-1D}    & \textbf{CL-CSEN-1D}   & \textbf{CSEN-1D}      & \textbf{CL-CSEN-1D}     \\ \hline
ARD $\downarrow$     & 0.3000 $\pm$ 0.013      & 0.2092 $\pm$ 0.005         & 0.3071 $\pm$ 0.011   & 0.2289 $\pm$ 0.004     & 0.2507 $\pm$ 0.006
     & 0.1978 $\pm$ 0.005       \\
SRD $\downarrow$     & 2.4741 $\pm$ 0.155      & 1.3051 $\pm$ 0.059         & 2.5457 $\pm$ 0.129   & 1.5757 $\pm$ 0.058     & 1.7831 $\pm$ 0.077     & 1.2107 $\pm$ 0.060        \\
RMSE $\downarrow$    & 6.3539 $\pm$ 0.029       & 4.4239 $\pm$ 0.024         & 6.5480 $\pm$ 0.049   & 4.8954 $\pm$ 0.026     & 5.5808 $\pm$ 0.037     & 4.1575 $\pm$ 0.027       \\
$\text{RMSE}_{\text{log}}$ $\downarrow$ & 0.5490 $\pm$ 	0.092       & 0.2885 $\pm$ 0.006         & 0.5717 $\pm$ 0.064   & 0.3083 $\pm$ 0.009     & 0.4283 $\pm$ 0.060     & 0.2660 $\pm$ 0.004       \\
$\delta < 1.25$ $\uparrow$     & 0.6182 $\pm$ 0.006     & 0.7765 $\pm$ 0.004         & 0.5957 $\pm$ 0.004   & 0.7364 $\pm$ 0.001     & 0.6762 $\pm$ 0.009     & 0.8087 $\pm$ 0.002       \\
$\delta < 1.25^2$ $\uparrow$     & 0.8492 $\pm$ 0.006      & 0.9297 $\pm$ 0.002          & 0.8375 $\pm$ 0.007    & 0.9111 $\pm$ 0.001     & 0.8845 $\pm$ 0.008     & 0.9414 $\pm$ 0.002       \\
$\delta < 1.25^3$ $\uparrow$     & 0.9251 $\pm$ 0.006      & 0.9672 $\pm$ 0.002          & 0.9199 $\pm$ 0.006    & 0.9602 $\pm$ 0.002     & 0.9432 $\pm$ 0.006    & 0.9716 $\pm$ 0.001       \\ \hline
\end{tabular}
}
\end{subtable}

\bigskip
\noindent
\begin{subtable}{1\textwidth}
\centering
\caption{The train:test splits are selected as approximately 1:17 proportion and the distance sensitivity (with quantization) is 1m.}
\resizebox{.8\linewidth}{!}{
\begin{tabular}{c|cc|cc|cc}
        $\phi(.)$ & \multicolumn{2}{c|}{DenseNet-121} & \multicolumn{2}{c|}{VGG19} & \multicolumn{2}{c}{ResNet-50} \\ \hline
        & \textbf{CSEN-1D}       & \textbf{CL-CSEN-1D}       & \textbf{CSEN-1D}    & \textbf{CL-CSEN-1D}   & \textbf{CSEN-1D}      & \textbf{CL-CSEN-1D}     \\ \hline
ARD $\downarrow$     & 0.3365 $\pm$ 0.023      & 0.3457 $\pm$ 0.018         & 0.3365 $\pm$ 0.033   & 0.3195 $\pm$ 0.012     & 0.2923 $\pm$ 0.017
     & 0.3608 $\pm$ 0.017       \\
SRD $\downarrow$     & 3.0839 $\pm$ 0.351      & 3.3374 $\pm$ 0.293         & 3.0596 $\pm$ 0.492   & 2.8948 $\pm$ 0.212     & 2.3585 $\pm$ 0.227     & 3.3650 $\pm$ 0.225        \\
RMSE $\downarrow$    & 7.0004 $\pm$ 0.207       & 6.5406 $\pm$ 0.144         & 7.0975 $\pm$ 0.293   & 6.6121 $\pm$ 0.209     & 6.2865 $\pm$ 0.234     & 6.3010 $\pm$ 0.079       \\
$\text{RMSE}_{\text{log}}$ $\downarrow$ & 0.7130 $\pm$ 0.193       & 0.3893 $\pm$ 0.007         & 0.8101 $\pm$ 0.201   & 0.4091 $\pm$ 0.054     & 0.6495 $\pm$ 0.210     & 0.3913 $\pm$ 0.011       \\
$\delta < 1.25$ $\uparrow$     & 0.5573 $\pm$ 0.012     & 0.6219 $\pm$ 0.005         & 0.5408 $\pm$ 0.015   & 0.5944 $\pm$ 0.008     & 0.6056 $\pm$ 0.014     & 0.6273 $\pm$ 0.006       \\
$\delta < 1.25^2$ $\uparrow$     & 0.8064 $\pm$ 0.016      & 0.8417 $\pm$ 0.005          & 0.7930 $\pm$ 0.014    & 0.8323 $\pm$ 0.004     & 0.8478 $\pm$ 0.012     & 0.8385 $\pm$ 0.007       \\
$\delta < 1.25^3$ $\uparrow$     & 0.9003 $\pm$ 0.015      & 0.9221 $\pm$ 0.004          & 0.8930 $\pm$ 0.013    & 0.9230 $\pm$ 0.004     & 0.9247 $\pm$ 0.013    & 0.9172 $\pm$ 0.006       \\ \hline
\end{tabular}
}
\end{subtable}
\label{1D-CSEN}
\end{table*}

\section{Conclusion}
\label{conclusion}

In this study, we first propose a novel CSEN-based distance estimation method using a single camera. CSENs were recently proposed to directly estimate support sets of a signal instead of the traditional approach, i.e., first reconstructing the sparse signal and applying a threshold. Using the modified CSENs for regression, we demonstrate that it is possible to utilize representative dictionaries for a regression task; and to the best of authors' knowledge, this makes the pioneer study in this domain. Hence, we introduce the term \textit{Representation-based Regression (RbR)} to reflect this fact. Moreover, utilizing the introduced representative dictionary design by collecting the samples with the same distances in the quantization level, the performance of the proposed distance estimators becomes class invariant unlike the several existing studies such as \cite{disnet,iccv}.

Finally, we propose a novel CSEN architecture in the CL-CSEN model by introducing the ability to fine-tune the proxy mapping matrix during the training procedure. Therefore, the proposed CL-CSEN method is a complete, one-to-one support estimator network in which the denoiser matrix $\mathbf{B}$ is directly connected to the convolutional layers using fully connected dense layers. Thus, it provides a superior distance estimation performance and efficient single-stage inference. Overall, it is observed that CSEN and CL-CSEN architectures significantly outperform the competing methods, SVR, CRC, and SRC. Finally, with their compact network models, we have shown that both CSEN and CL-CSEN are able to learn with a limited number of annotated data, e.g., with less than 13$\%$ annotated data used in the training to demonstrate this competence.

\ifCLASSOPTIONcaptionsoff
  \newpage
\fi



%

\bibliographystyle{IEEEtran}
\bibliography{IEEEtran}




\end{document}